\documentclass{article}

\usepackage{microtype}
\usepackage{graphicx}
\usepackage[]{caption}
\usepackage{subcaption}
\usepackage{booktabs} 
\usepackage{amsmath,amssymb, amsthm}
\usepackage[usenames, dvipsnames, table]{xcolor}
\usepackage{hyperref}
\usepackage{cleveref}
\usepackage{float}
\usepackage{algorithm}
\usepackage{algpseudocode}

\usepackage[markup=bfit, deletedmarkup=sout, authormarkup=superscript]{changes}
\definechangesauthor[name={kw}, color=MidnightBlue]{KW}
\definechangesauthor[name={td}, color=red]{TODO}
\definechangesauthor[name={ms}, color=orange]{MS}
\definechangesauthor[name={ee}, color=green]{EE}

\newtheorem{theorem}{Theorem}

\usepackage[preprint]{corl_2021}

\usepackage{xifthen}  

\newcommand{\policy}[1][]{
    \ifthenelse{\isempty{#1}}
        {\pi(a|s)}  
        {\pi(a_{#1}|s_{#1})}  
}

\newcommand{\reward}[1][]{
    \ifthenelse{\isempty{#1}}
        {R(a, s)}  
        {R(a_{#1}, s_{#1})}  
}

\newcommand{\transition}[1][]{
    \ifthenelse{\isempty{#1}}
        {T(s' | s, a)}  
        {T(s_{#1+1} | s_{#1}, a_{#1})}  
}

\title{SHIRO: Soft Hierarchical Reinforcement Learning}

\author{
  Kandai Watanabe\\
  Department of Computer Sciences\\
  University of Colorado Boulder\\ 
  United States\\
  \texttt{kandai.watanabe@colorado.edu} \\
  \And
  Mathew Strong \\
  Department of Computer Sciences\\
  University of Colorado Boulder\\ 
  United States\\
  \texttt{mathew.strong@colorado.edu} \\
  \And
  Omer Eldar \\
  Department of Computer Sciences\\
  University of Colorado Boulder\\ 
  United States\\
  \texttt{omer.eldar@colorado.edu} \\
}

\begin{document}
\maketitle

\begin{abstract}

Hierarchical Reinforcement Learning (HRL) algorithms have been demonstrated to perform well on high-dimensional decision making and robotic control tasks. However, because they solely optimize for rewards, the agent tends to search the same space redundantly. This problem reduces the speed of learning and achieved reward.
In this work, we present an Off-Policy HRL algorithm that maximizes entropy for efficient exploration. The algorithm learns a temporally abstracted low-level policy and is able to explore broadly through the addition of entropy to the high-level.
The novelty of this work is the theoretical motivation of adding entropy to the RL objective in the HRL setting. We empirically show that the entropy can be added to both levels if the Kullback-Leibler (KL) divergence between consecutive updates of the low-level policy is sufficiently small.
We performed an ablative study to analyze the effects of entropy on hierarchy, in which adding entropy to high-level emerged as the most desirable configuration.
Furthermore, a higher temperature in the low-level leads to Q-value overestimation and increases the stochasticity of the environment that the high-level operates on, making learning more challenging. Our method, SHIRO, surpasses state-of-the-art performance on a range of simulated robotic control benchmark tasks and requires minimal tuning.

\end{abstract}

\keywords{Off-Policy, Hierarchical, Reinforcement Learning, Maximum Entropy} 

\section{Introduction}
\label{introduction}


Deep Reinforcement Learning (RL) is a promising approach for robots to learn complex policies; e.g., manipulators \cite{gu:2017:DRL4Manipulation}, quadrupedal robots \cite{yang:2020:HIRO4Legged, jain:2019:HRL4Legged} and humanoids \cite{duan:2016:benchmarking, peng:2018:deepmimic, peng:2019:mcp}. Learning complex policies is made possible with the help of strong function approximators such as neural networks. This enables an agent to learn a policy in both a high dimensional state and action space \cite{silver:2016:alphago}.
Deep RL methods are well-suited for learning one atomic skill; e.g., a manipulator reaching an object or a quadruped walking upright. While very powerful, such methods struggle to learn compositional, long-horizon skills, such as utilizing locomotion for solving a maze. To solve such tasks, a better structure and stronger exploration strategies are required.

To this end, Hierarchical Reinforcement Learning (HRL) has been empirically shown to work well on these tasks by making use of the temporal abstraction generated from its hierarchical structure \cite{nachum:2018:hiro, levy:2018:multi-level-hindsight, jiang:2019:language-as-abstraction, nachum:2018:nearoptimal}.
In HRL, policies are stacked into multiple levels to construct a hierarchy, where the low-level agent receives information from the one(s) above it. This allows for the decomposition of a long-horizon task into smaller, more easily solvable problems. The theory of HRL has been well-studied over the years, but has lacked a generalized method that can train an agent efficiently in a complex, continuous space \cite{dayan:1993:feudal, parr:1998:hierarchiesofmachines, sutton:1999:betweenmdp, dietterich:2000:MAXQ, nachum2019does}.

One potential solution is the state-of-the-art algorithm named HIerarchical Reinforcement Learning With Off-Policy Correction (HIRO) \cite{nachum:2018:hiro}. It proposes to stack two off-policy Deep RL agents and to relabel past experiences in the high-level policy's replay buffer to correct the changes in the low-level policy. This allows training both levels simultaneously using an off-policy training method, which speeds up the training.
Accordingly, this method has been extended to new algorithms \cite{nachum:2018:nearoptimal, jiang:2019:language-as-abstraction, jain:2019:HRL4Legged, zhang2020generating, lilearning}.
However, because the agent solely optimizes for reward, it over-commits to actions it believes are optimal, leading to poor exploration. This leads to inefficient training and slower convergence.


We present an off-policy hierarchical reinforcement learning algorithm (SHIRO: Soft Hierarchical Reinforcement Learning with Off-Policy Correction).
We propose adding entropy maximization to the objective \cite{haarnoja:2018:sac} of both levels of the agent to improve the training efficiency of HIRO. The addition of entropy in the RL objective has demonstrated an improvement of efficiency in exploration \cite{ziebart2008maximum, toussaint2009robot, rawlik2012stochastic, fox2016taming, haarnoja2017reinforcement, haarnoja:2018:sac, levine2018reinforcement}.
Additionally, we show that a relatively small Kullback-Leibler (KL) divergence in the low-level agent over time guarantees that the low-level agent can be viewed as part of the environment. This allows us to train the high-level agent as if it was a single agent and applies the soft policy iteration theorem in SAC to our work.
We empirically demonstrate more efficient learning with the addition of entropy to the RL objective. Our method produces superior results in data efficiency and training time to the base implementation of HIRO on HRL benchmark environments for robotic locomotion (AntMaze, AntFall, and AntPush).
Finally, we conduct an ablative study, where we apply different combinations of entropy, and experiment with making the entropy temperature a learned parameter. The results of the ablative study are used to produce the best possible method.

\section{Related Work} \label{sec:RelatedWork}

The problem of finding a suitable hierarchical structure to solve a long-horizon task has been widely explored in the RL community for many years \cite{dayan:1993:feudal, parr:1998:hierarchiesofmachines, sutton:1999:betweenmdp, dietterich:2000:MAXQ}. General structures are a tree-like structure where multiple policies are stacked like a tree and a tower-like structure where single policies are stacked on top of each other.
Classic tree-like works train a high-level policy, given a set of hand-engineered low-level policies \cite{stolle:2002:learningoptions, mannor:2004:dynamic, singh:2005:intrinsically}.
One popular framework in a tree-like structure is the Options framework \cite{sutton:1999:betweenmdp, precup:2000:temporal}.
This algorithm learns to pick an option and follows the selected option policy until it meets its termination condition. We can view the high-level policy as a switch over options. However, it requires prior knowledge for designing options. To this end, the option-critic \cite{bacon:2017:option} proposes a method to train the high-level policy jointly. Recent work on the Options framework utilizes regularizers to learn multiple sub-policies \cite{bacon:2017:option, harb:2018:waiting, mnih:2016:strategic}.

Recent works focus on designing rewards to learn sub-policies more effectively. One choice is to learn both high-level and low-level policies from final task rewards end-to-end \cite{frans:2018:meta, sigaud:2019:policy}, including the aforementioned option-critic frameworks \cite{bacon:2017:option, harb:2018:waiting}.
Another major stream is to provide auxiliary rewards to low-level policies to foster the learning \cite{singh:2005:intrinsically, heess:2016:learning, kulkarni:2016:hierarchical, tessler:2017:deep, florensa:2017:stochastic}, such as hand-engineered low-level reward based on domain knowledge \cite{konidaris:2007:building, heess:2016:learning, kulkarni:2016:hierarchical, tessler:2017:deep}, mutual information for more efficient exploration \cite{daniel:2012:hierarchical, florensa:2017:stochastic} and goal-conditioned rewards \cite{dayan:1993:feudal, schaul:2015:universal, andrychowicz:2017:hindsight, levy:2018:multi-level-hindsight, vezhnevets:2017:feudal,nachum:2018:hiro,nachum:2018:nearoptimal}.

The tower-like structure, such as the FeUdal framework \cite{dayan:1993:feudal,vezhnevets:2017:feudal} and HIRO algorithms \cite{nachum:2018:hiro}, generally takes a form of goal-conditioned rewards, so that the high-level policy acts as a Manager/Planner to lead the low-level policy to work for/reach the sub-goal that the high-level policy provided. The benefit of these methods is that the representation is generic and does not require hand-engineering specific to each domain. Especially, HIRO \cite{nachum:2018:hiro} uses the state in its raw form to construct a parameterized reward and \cite{nachum:2018:nearoptimal,jiang:2019:language-as-abstraction} extend the sub-goal to latent spaces. This goal-conditioned framework is also extended to the real robotic domain \cite{jain:2019:HRL4Legged, nachum2020multi} and extensively explored outside of HRL \cite{mahadevan:2007:proto, sutton:2011:horde, schaul:2015:universal,andrychowicz:2017:hindsight, pong:2018:temporal}. Problems of these algorithms are also targeted as in \cite{li2019sub, hejna2020hierarchically, zhang2020hierarchical, li2019hierarchical}, indicating that there are numerous avenues for improvement. 

Furthermore, while very powerful, HIRO searches the same states redundantly, leading to inefficiency. Our algorithm extends the work to maximize entropy in addition to the original RL objective, that is shown to work well in other works \cite{ziebart2008maximum, toussaint2009robot, rawlik2012stochastic, fox2016taming, haarnoja2017reinforcement, haarnoja:2018:sac, levine2018reinforcement}, including finding diverse goals to speed up learning \cite{zhao2019maximum}. It is also applied in HRL \cite{haarnoja2018latent} and also in the goal-conditioned setting \cite{azarafrooz2019hierarchical, tang2021novel}. While \cite{azarafrooz2019hierarchical, tang2021novel} are similar to our work, the intrinsic reward for the low-level controller must be hand-engineered and the controller is pre-trained to obtain enough abstraction before the joint training. \cite{lilearning} is the closest to our work where they use SAC on both levels, but they don't explain on what condition entropy works well. While maximizing entropy is effective, too high entropy leads to poor performance. Regarding the problems of sub-goals, \cite{zhang2020generating} added a constraint to avoid from generating unreachable sub-goals, and \cite{zhao2019maximum} minimizes changes in the high-level policy which facilitates learning. Unlike other methods, our work adds entropy to either or both levels and explains on what condition adding entropy works and how to best utilize entropy.


\section{Preliminaries}
\label{sec:Preliminaries}

\subsection{Goal Conditioned Reinforcement Learning} \label{subsec:GoalConditionedRL}
Goal-Conditioned Reinforcement Learning (RL) extends the definition of the RL problem by conditioning its policy and reward function on goals. To define formally, a Markov Decision Process (MDP) is augmented with a set of goals $\mathcal{G}$, and is called a \textit{Universal} MDP (UMDP). A UMDP is a tuple containing $(\mathcal{S}, \mathcal{G}, \mathcal{A}, P_T, R)$, where $\mathcal{S}$ is the state space, $\mathcal{A}$ is the action space, $P_T: \mathcal{S} \times \mathcal{A} \times \mathcal{S} \rightarrow [0, 1]$ is the transition probability and $R: \mathcal{S} \times \mathcal{G} \times \mathcal{A} \rightarrow [r_{min}, r_{max}]$ is the reward function mapping state, goal and action pair to a scalar value reward between $r_{min}$ and $r_{max}$.  The reward function is conditioned on a goal to represent how well the agent is approaching the goal. To approach each goal in a different manner, the agent's policy $a_t = \pi(a_t | s_t, g)$ is augmented to condition on the goal. At the start of every episode, a goal $g$ is sampled from a set of goals $\mathcal{G}$ by the distribution $P_g(g)$. At every time step $t$, the environment generates state $s_t \in S$. Then, the agent takes action $a_t \in \mathcal{A}$ sampled from its policy $\pi(a_t|s_t, g)$, transitions to the next state $s_{t+1}$ with probability $P_T(s_{t+1}|s_t, a_t)$, and receives a reward $R_t=R(s_t, a_t, g)$. The probability distribution over the samples generated by the policy $\pi$ is denoted as $\rho^\pi$. The objective is to find the optimal policy $\pi_{\phi}$ with parameters $\phi$ that maximizes the expected reward $\mathbb{E}_{(s_t, a_t)\sim \rho^\pi, \ g \sim P_g}[\Sigma_t \gamma^t R(s_t, a_t, g)]$.

\begin{figure}
\centering
\includegraphics[width=\textwidth]{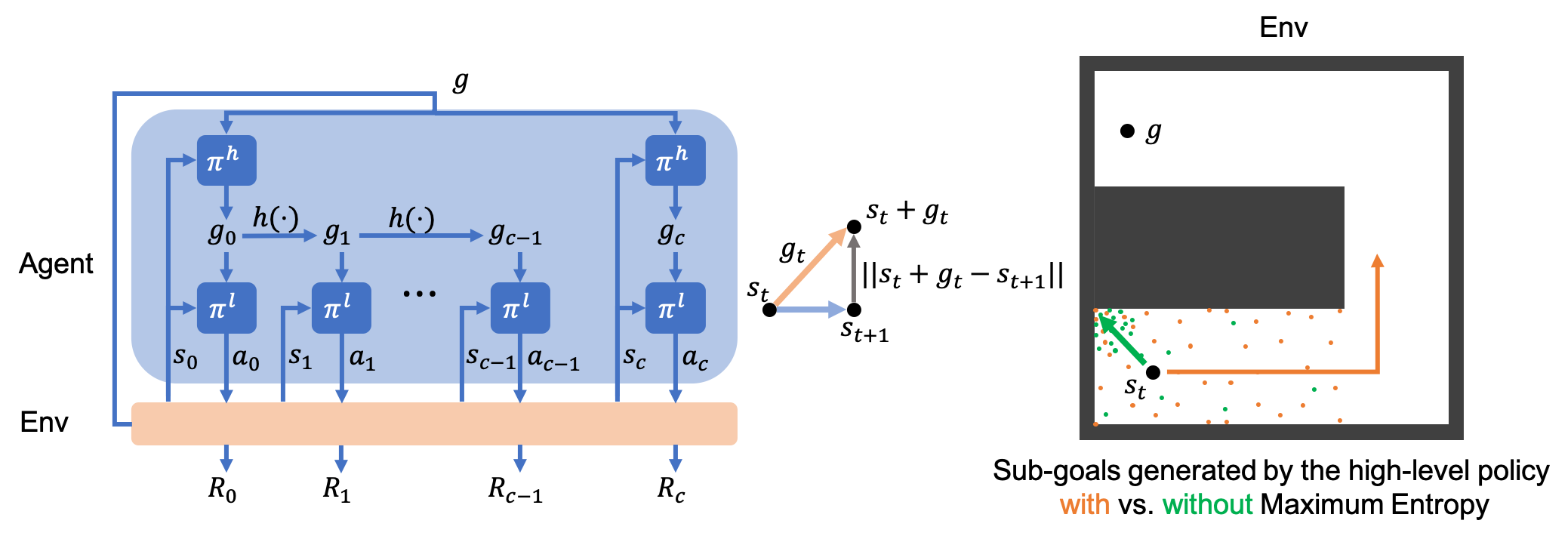}
\caption{The left figure is a schematic of how the agent interacts with the environment. The high-level policy generates a sub-goal to guide the low-level policy in its direction as shown in the center figure. The right figure is the depiction of the high-level policy generating sub-goals at time $t$ in AntMaze (from \Cref{fig:env}). The entropy maximization helps the policies to diversify their sub-goals/actions for better exploration, allowing to get out of the local optima, faster convergence, and improved training of the low-level policy.}
\label{fig:hiro_schematic}
\end{figure}

\subsection{Goal Conditioned Hierarchical Reinforcement Learning} \label{subsec:HIRO}

HRL is an RL approach that leverages the hierarchical structure in an agent's model to decompose the original problem into sub-problems. This allows the algorithm to learn within the space of each sub-problem, and to learn a policy that can be shared across the sub-problems. HIerarchical Reinforcement learning with Off-policy correction (HIRO) \cite{nachum:2018:hiro} is a potential solution to this problem that makes use of goal-conditioned policies. This algorithm successfully accomplishes continuous complex tasks and is widely used \cite{jiang:2019:language-as-abstraction, beyret:2019:dot2dot, nachum:2018:nearoptimal}. Our work is based on their work, and is depicted in \Cref{fig:hiro_schematic}. In the HIRO algorithm, the agent consists of a high-level policy $\pi^h(g_t | s_t, g)$  and a low-level policy $\pi^l(a_t | s_t, g_t)$, where $g_t$ is a sub-goal sampled from the high-level policy. At every $c$ steps, the high-level policy computes a sub-goal for the low-level policy given the current state $s_t$ and final goal $g$. The low-level policy is conditioned on the current state $s_t$ and the sub-goal $g_t$ to generate an action $a_t$. For the rest of the steps, the sub-goal is advanced by the goal transition model $h(s_t, g_t, s_{t+1}) = s_t + g_t - s_{t+1}$ based on the current and next state. The intrinsic reward is defined as $r(s_t, g_t, a_t, s_{t+1}) = -|| s_t + g_t - s_{t+1} ||_2$ to reward the low-level policy, while the environment reward is for the high-level policy.
To train the policies, the experiences are stored into replay buffers and apply off-policy training individually. HIRO addresses the non-stationary problem by relabeling past replay experiences with new goals that would prompt the current low-level policy to behave in the same way as the low-level policy in the past.

\subsection{Soft Actor Critic}\label{subsec:SoftActorCritic}
Recent advancements in off-policy RL have enabled policies to be trained on a range of challenging domains, from games to robotic control.
Off-policy algorithms such as Q-learning, DDPG, and TD3 \cite{lillicrap2016continuous, fujimoto2018addressing} learn a Q-function parameterized by $\theta$ that estimates the expected total discounted reward given a state and an action.
The Q function can be computed iteratively as $Q^\pi(s_t, a_t) = \mathbb{E}_{s_{t+1} \sim P_T}[R(s_t, a_t)+\gamma \mathbb{E}_{a_{t+1} \sim \pi} (Q^{\pi}(s_{t+1}, a_{t+1}))]$, such that it minimizes the Bellman error over all sampled transitions.
The deterministic policy can then be updated through the deterministic policy gradient \cite{lillicrap2016continuous}.
However, the standard formulation faces two major challenges; high sample complexity and brittle convergence. To this end, Soft Actor-Critic (SAC) \cite{haarnoja:2018:sac} proposes to alter the original objective to include entropy maximization, i.e., $J(\pi) = \sum_{t=0}^{T} \mathbb{E}_{(s_t, a_t) \sim \rho^\pi} [R(s_t, a_t) + \alpha \mathcal{H}(\pi(\cdot | s_t))]$,
where $\alpha$ is a temperature parameter that
tunes the stochasticity of the optimal policy and $\mathcal{H}$ represents the entropy of the policy. The Q-function and value function can be computed iteratively by:
\begin{align}
Q^{\pi}(s_t, a_t) &= R(s_t, a_t) + \gamma \mathbb{E}_{s_{t+1}\sim P_T} \Big[ V^{\pi}(s_{t+1}) \Big] \label{eq:qFunc}\\
V^{\pi}(s_t) &= \mathbb{E}_{a_t \sim \pi} [ Q^{\pi}(s_t, a_t) - \alpha \log \pi (a_t | s_t)] \label{eq:valueFunc}
\end{align}
The last term adds a larger value to the original Q function if the probability of taking action $a$ is small so that it explores actions that are less likely to occur. By adding entropy, it improves exploration, allowing for searching sub-optimal actions that may potentially lead to better rewards. The policy can then be trained to minimize the KL divergence $J_{\pi}(\phi) = \mathbb{E}_{s_t \sim \rho^\pi} [\rm{KL} (\pi_{\phi}(\cdot|s_t) \| exp(Q^{\pi_{\rm{old}}}_{\theta}(s_t, \cdot)) / Z^{\pi_{\rm{old}}}_{\theta})]$ where $Z^{\pi_{\rm{old}}}$ is a partition function that normalizes the distribution.


\section{Soft Hierarchical Reinforcement Learning} \label{sec:SoftHRL}

While HIRO is very effective, it searches the same areas redundantly (see \Cref{sec:Experiments}), which contributes to data inefficiency, slowing down learning.
In our algorithm, we propose to add entropy to the high-level policy to improve exploration. Furthermore, we also show how it can be best added to both levels. In Maximum Entropy RL, e.g., the Soft Actor Critic algorithm, the objective is to maximize the expectation of accumulated rewards and the policy entropy for a single agent. Similarly, in SHIRO, we also maximize entropy as follows
\begin{equation}
    J(\pi^h) = \sum\limits_{t\in T_c} \mathbb{E}_{(s_t, g_t) \sim\rho^{\pi^{h}}} \Bigl[ R^{abs}(s_t, g_t) + \alpha \mathcal{H}(\pi^{h}(\cdot | s_t, g)) \Bigr]
    \label{eq:MaxEntObjForHRL}
\end{equation}
where we define an Abstracted Reward Function $R^{abs}(s_t, g_t) = \sum_{t'=t}^{t+c} R(s_{t'}, \pi^l(a_{t'}|s_{t'}, g_t))$. In the case of goal-conditioned HRL, the high-level agent provides a new sub-goal every $c$ steps, where $T_c$ is a set of time step $t$ at every $c$ steps, defined as $T_c = \{t | t = c \cdot n, n=0, 1,2, ... \}$. By introducing such a function, the objective can become analogous to the standard reinforcement learning objective if we assume that the low-level policy is a part of the environment.
Given this assumption, in order to transform the HRL problem into one that can be applied to standard RL, we introduce another function called the Abstracted Transition Function $P_T^{abs}(s_{t+c} | s_t, g_t)$; i.e., a transition function from a state $s_t$ to a state $s_{t+c}$, given a sub-goal $g_t$. With the abstraction, the probability of a trajectory $\tau$ can be rewritten as follows, where the original probability for the single policy is given as a multiplication of probabilities of state and action sequences along the trajectory, $p(\tau) = p(s_0) \prod_{t \in T_c} \Big[\pi^h(g_t|s_t, g) P_T^{abs}(s_{t+c}|s_t, g_t) \Big]$.
Given that the low-level policy gets updated, which induces a certain degree of stochasticity, we can view this Abstracted Transition Function as a representation of the stochastic environment.
This is exactly the same problem as in the standard Reinforcement Learning objective for the high-level policy, as long as the Abstracted Transition function does not change drastically. Formally, we can bound the change in the Abstracted Transition function by the following theorem:

\begin{theorem}
Let $\pi^l_{\phi}(a_t|s_t, g_t)$ be the low-level policy before the parameter update and  $\pi^l_{\phi^{\prime}}(a_t|s_t, g_t)$ be the updated low-level policy after $c$ steps. If two policies are \textbf{close} , i.e., $| \pi^l_{\phi^{\prime}} (a_t | s_t, g_t) - \pi^l_{\phi} (a_t | s_t, g_t) | \leq \epsilon$ for all $s_t$, then the change in the abstracted transition functions is bounded by a small value $|P_{T \pi^l_{\phi^{\prime}}}^{abs}(s_{t+c} | s_t, g_t) - P_{T \pi^l_{\phi}}^{abs}(s_{t+c} | s_t, g_t)| \leq 2 \epsilon c$, where $\epsilon \in [0, 1]$.
\label{theorem:KLbound}
\end{theorem}
See the proof in \Cref{sec:Proofs}.

\subsection{Addition of Entropy to Each Policy}
\textbf{High-Level Policy:} If the changes in the parameters are small, and if we can assume the low-level policy as a part of the environment, we can view the HRL problem in the same way as in the original RL problem, shown in \Cref{eq:MaxEntObjForHRL}. Then, we can derive the exact same theorems as in SAC \cite{haarnoja:2018:sac} for the high-level policy. For a deterministic high-level, we can apply the theorems of standard policy improvement and iteration. The addition of entropy to high-level is desirable for improved hierarchical exploration because it will generate more diverse sub-goals that facilitate learning of the low-level, leading to a better performing low-level agent to reach those broad sub-goals.



\textbf{Low-Level Policy:}
Simultaneously (or by itself), we can add entropy to the low-level policy according to \Cref{theorem:KLbound}.
To make the total variation divergence of policies between the parameter update small, we can use Pinsker's inequality, i.e., $|\pi^l_{\phi'}(a_t | s_t, g_t) - \pi^l_{\phi} (a_t | s_t, g_t)| \leq \sqrt{\frac{1}{2} \rm{KL} (\pi^l_{\phi'}(a_t | s_t, g_t) || \pi^l_{\phi} (a_t | s_t, g_t)) }$.
This means that if the KL divergence between parameter updates is kept low, we can bound the changes in the Abstract Transition function and allow the low-level policy to be part of the environment. 
In this paper, we empirically show that the high-level and both-levels agents can perform better than others, and we will verify that the KL divergence is sufficiently small during training in \Cref{sec:Experiments}.
Moreover, the temperature parameter should not be too large to avoid the second term in \Cref{eq:valueFunc} to be large, which induces an incorrect estimation of the value function. This induces overly high values on actions that are unlikely to be taken, messing up the estimation of the low-level policy.

\subsection{Learning Temperature Parameter}
Given that our proof allows to view the high-level agent as a single SAC agent, this motivates the use of learned entropy on the high-level. We apply the learned temperature method of \cite{haarnoja2018soft} into the high-level and low-level agent. The objective for learning entropy temperature is given by $\arg \min_{\alpha} \mathbb{E}_{a_t \sim {\pi_{t}^*}} \bigl[ - \alpha \log {\pi_{t}^*} (a_t | s_t, g_t) - \alpha \mathcal{\bar{H}} \bigr]$,
where $\bar{\mathcal{H}}$ is the minimum desired entropy, and $\pi_{t}^*$ is the policy at timestep $t$. The expression is a measure of the difference between the current policy's entropy and the minimum desired entropy. A lower value of $\alpha$ will minimize and scale down this difference. With respect to the high-level, we assign a high temperature at the beginning of training to speed up learning, but allowing the temperature to decrease once the agent has learned. Furthermore, this is sufficient as the low-level agent will have been sufficiently trained on diverse data initially; lowering entropy does not pose any major concern.
Similarly, we can also learn the temperature parameter $\alpha$ of the low-level agent, in which we desire a low entropy that won't increase significantly.

\section{Experiments} \label{sec:Experiments}

\begin{figure*}[t]
    \centering
    \begin{subfigure}[b]{0.3\textwidth}
        \centering
        \includegraphics[width=\textwidth]{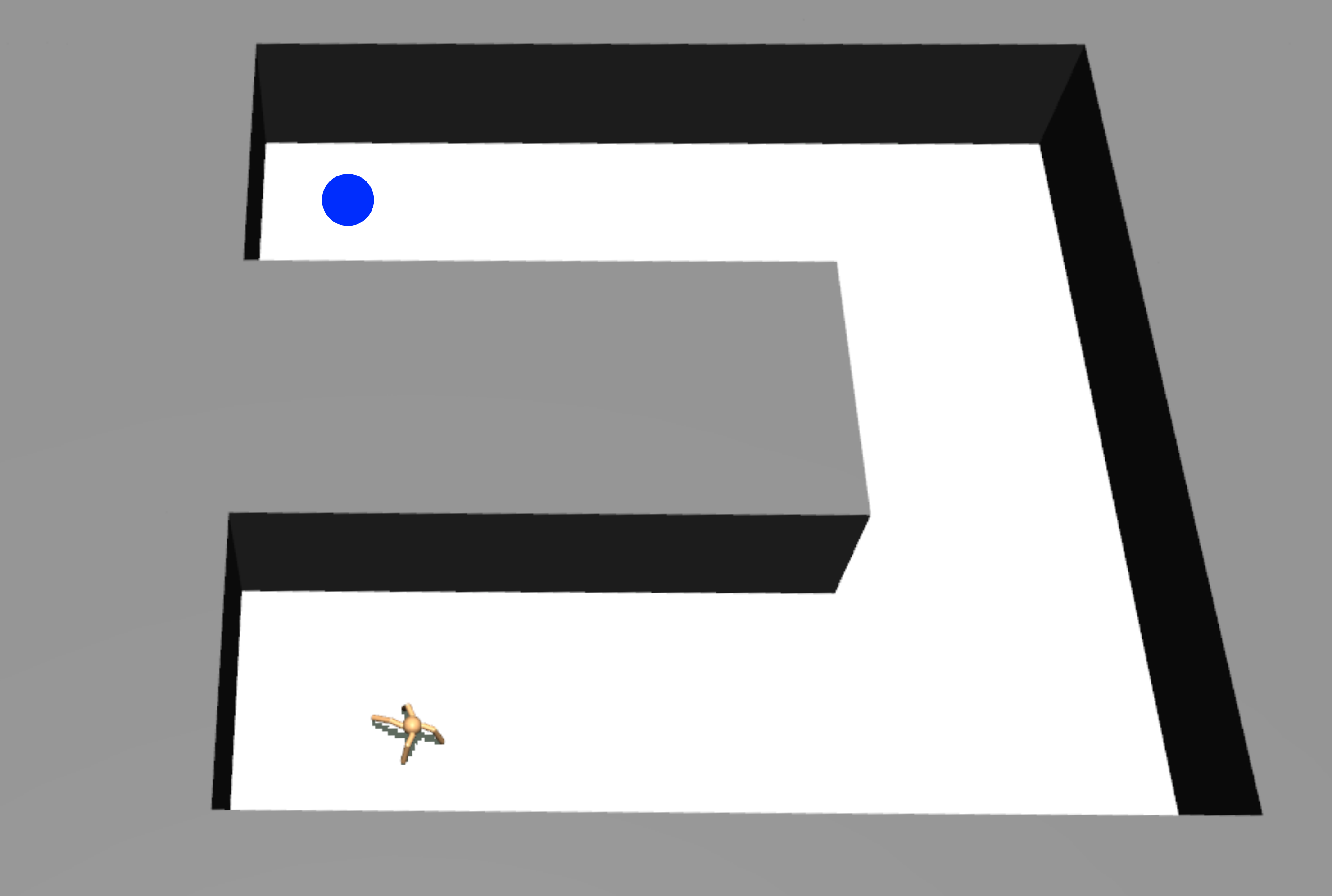}
        \caption{AntMaze}
    \end{subfigure}
    \hfill
    \begin{subfigure}[b]{0.3\textwidth}
        \includegraphics[width=\textwidth]{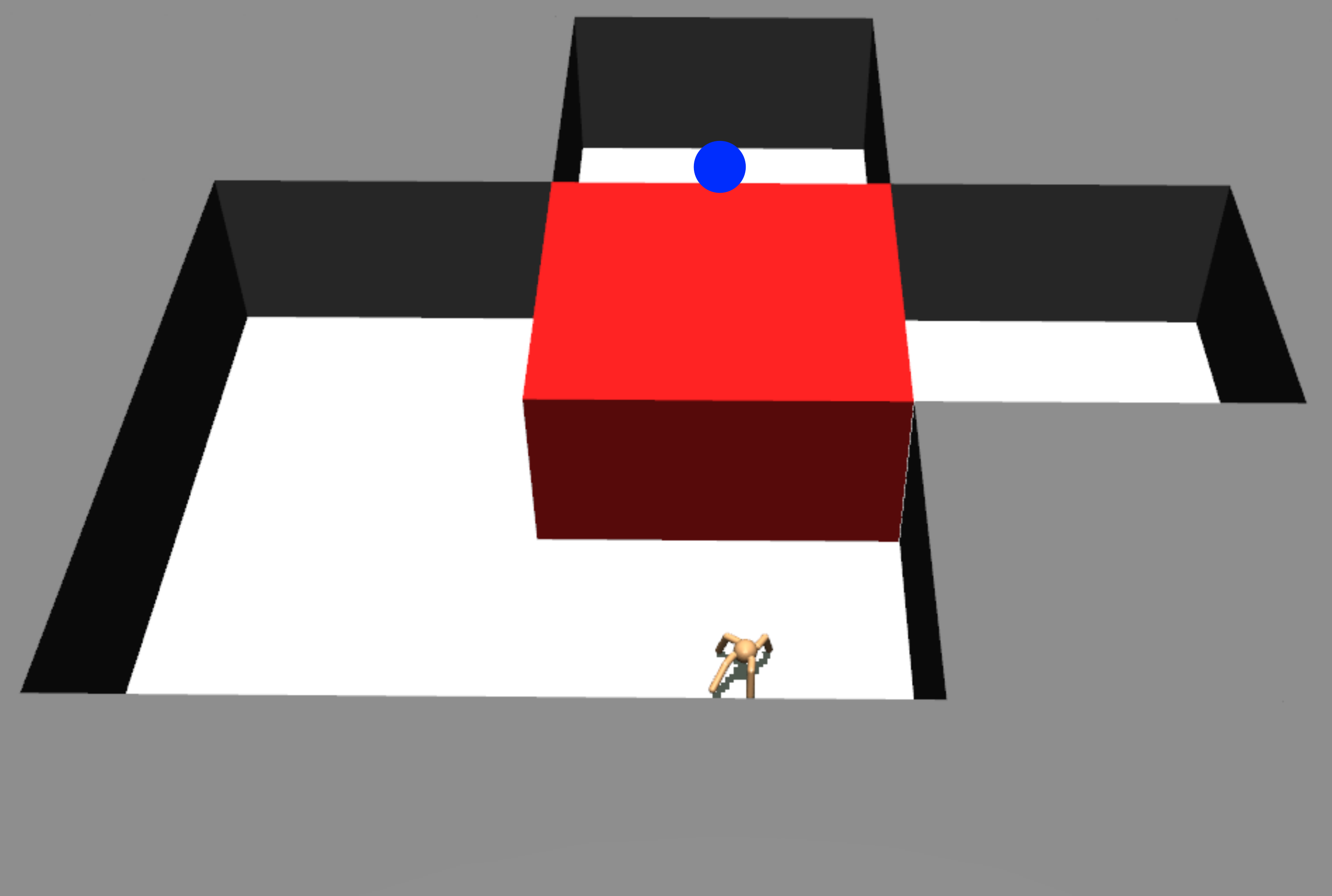}
        \caption{AntPush}
    \end{subfigure}
    \hfill
    \begin{subfigure}[b]{0.3\textwidth}
        \includegraphics[width=\textwidth]{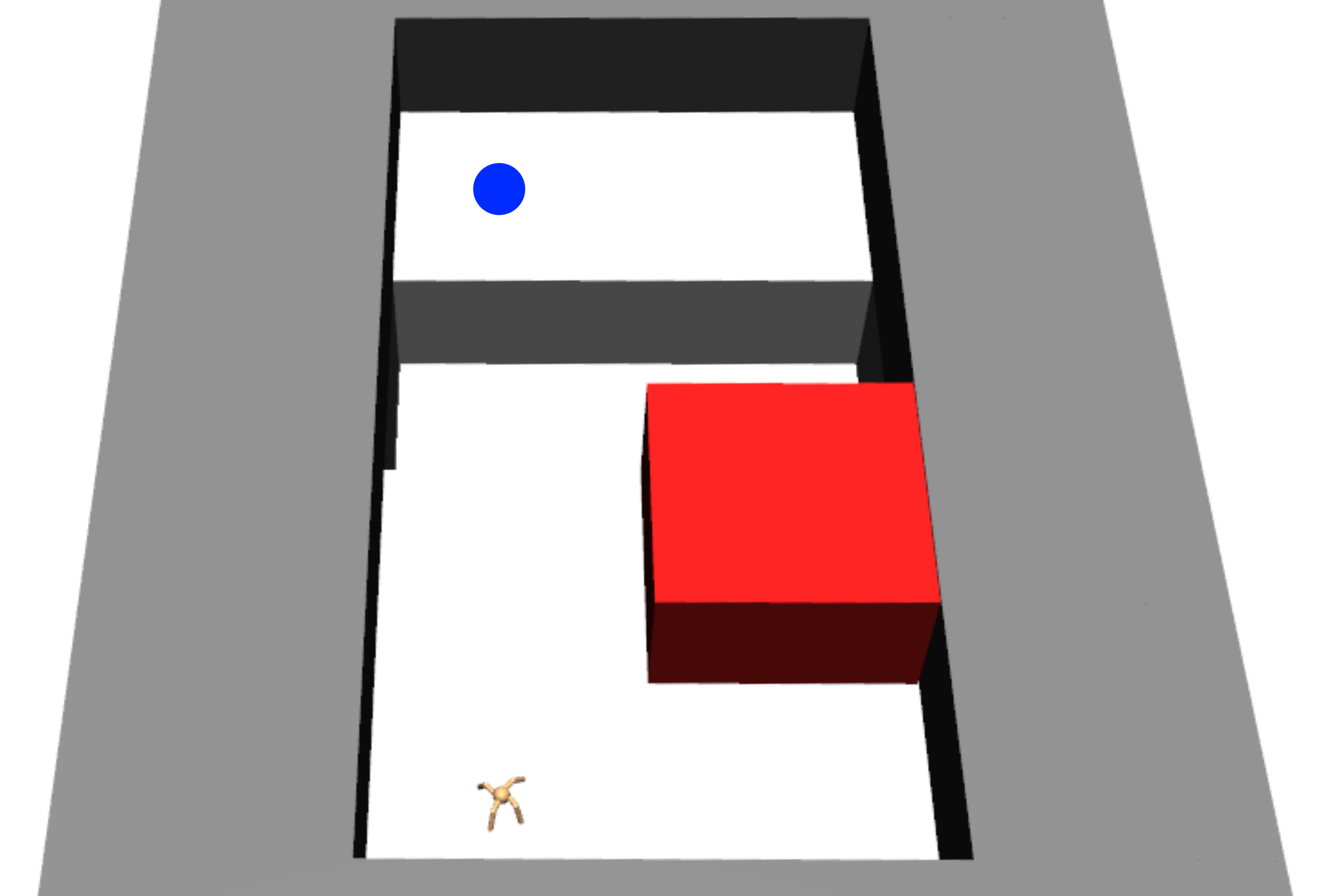}
        \caption{AntFall}
    \end{subfigure}
    \caption{Three environments used in the experiments. In each environment, the ant starts at the initial location and aims to reach the goal colored in blue.} 
    \label{fig:env}
\end{figure*}

We conducted two experiments to demonstrate the performance of our algorithms. The first experiment compares our best method to HIRO in benchmark environments. The second experiment is an ablative analysis of our algorithm. In this study, entropy is added to either or both of the levels, as well as with learning and not learning the temperature to analyze the behavior of the algorithm. These experiments demonstrate the power of the maximum entropy hierarchy with a simulated robot in MuJoCo.

\begin{figure*}[t]
    \centering
    \begin{subfigure}[b]{0.3\textwidth}
        \includegraphics[width=\textwidth]{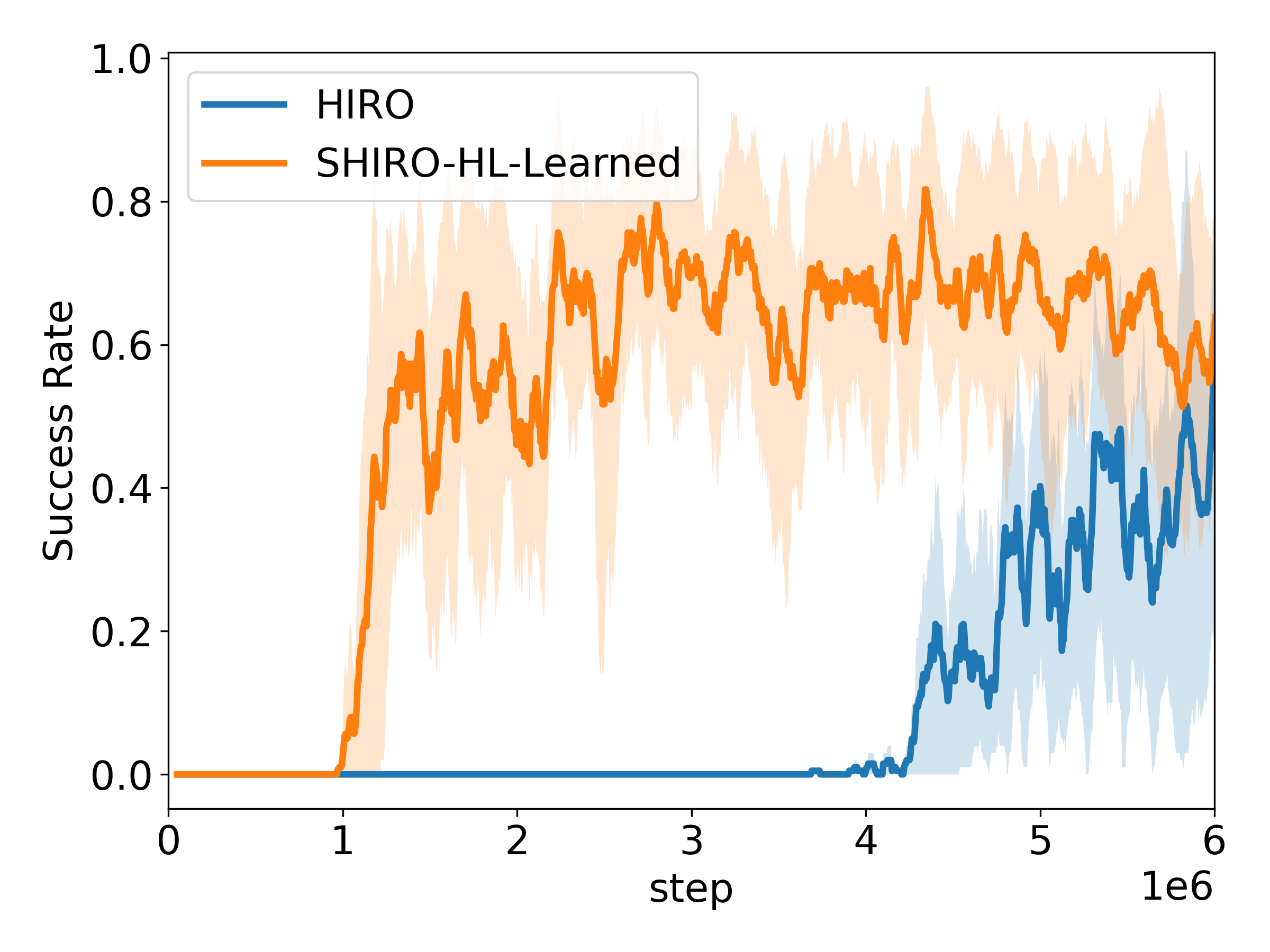}
        \caption{AntMaze}
    \end{subfigure}
    \hfill
    \begin{subfigure}[b]{0.3\textwidth}
        \includegraphics[width=\textwidth]{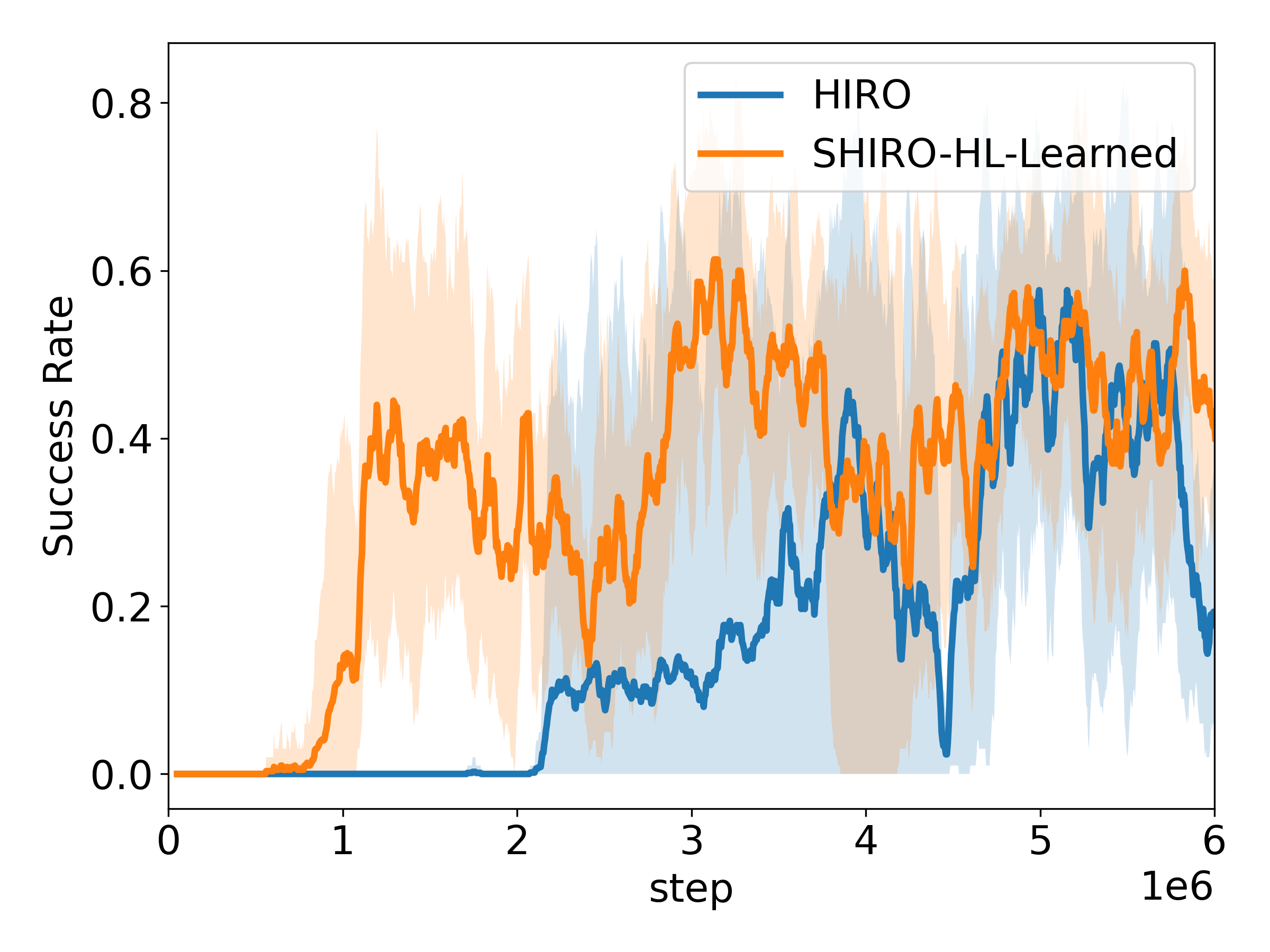}
        \caption{AntPush}
    \end{subfigure}
    \hfill
    \begin{subfigure}[b]{0.3\textwidth}
        \includegraphics[width=\textwidth]{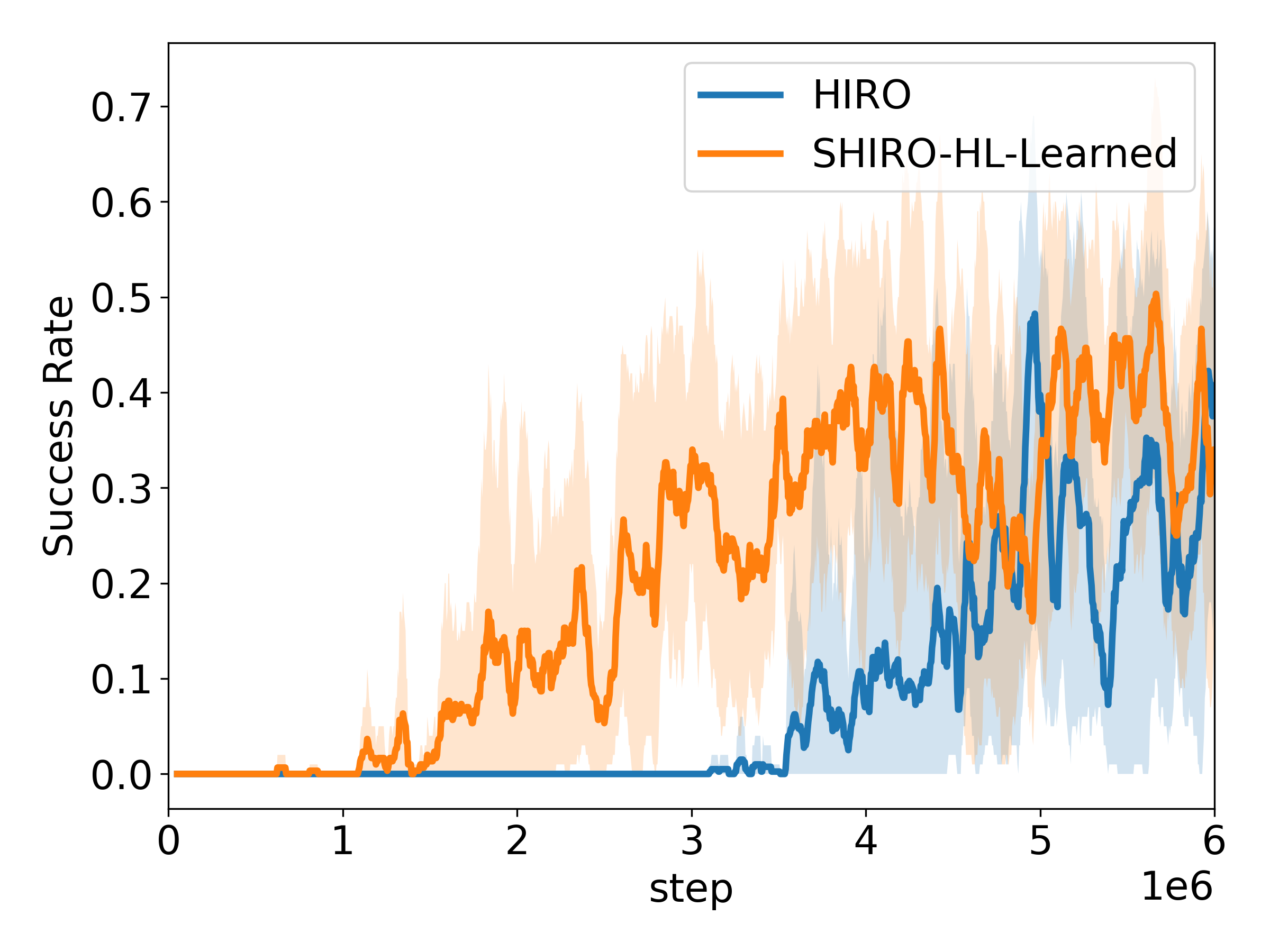}
        \caption{AntFall}
    \end{subfigure}
    \caption{Success rate across different environments comparing SHIRO HL-Learned to HIRO. The darker lines represent the mean and the shaded areas represent the minimum and the maximum values across 3 different seeds.}
    \label{fig:main_result}
\end{figure*}

Concretely, the agent is modeled as in \Cref{fig:hiro_schematic}. The structures and the hyperparameters of the policies and Q-functions in the deterministic case are exactly the same as in \cite{nachum:2018:hiro} to compare against HIRO fairly. High-level and low-level agents are trained simultaneously, while the high-level agent is updated every $c=10$ environment steps whereas the low-level agent is updated every environment step. The training of each agent is analogous to training a single agent. They can be trained independently using the samples from the replay buffer that are simultaneously relabeled by the off-policy correction method during the training. Most importantly, the entropy is defined and added to the original objective as in \Cref{eq:MaxEntObjForHRL}. The entropy is added to either or both of the policies, and they are compared against HIRO, as in \Cref{fig:main_result}. When we maximize entropy on either layer, we use the parameters according to Haarnoja's (\cite{haarnoja:2018:sac}) SAC. We call each of the variants, SHIRO HL (high-level entropy), SHIRO LL (low-level entropy), and BL (entropy at both levels) and used a temperature of $1.0$ for high-level policy and $0.1$ for the low-level policy across all environments. Other hyperparameters were carried over from SAC. We also used the same value as the initial temperatures for agents with temperature learning (SHIRO HL-Learned, SHIRO LL-Learned, SHIRO BL-Learned). All of the implementations is built on top of PFRL in PyTorch \cite{JMLR:v22:20-376}.

\subsection{Comparative Study}
In this study, we compared our \textit{best} performing method, SHIRO-HL Learned, against HIRO \cite{nachum:2018:hiro} to demonstrate the effectiveness of entropy maximization formulation. HIRO is known to perform better than other HRL methods such as FeUdalNet\cite{vezhnevets:2017:feudal} and Option-Critic\cite{bacon:2017:option} on the same tasks \cite{nachum:2018:hiro}. Furthermore, we also compared our methods to the vanilla SAC (no hierarchy). 

\begin{figure}[t]
    \centering
    \begin{subfigure}[b]{.45\linewidth}
        \includegraphics[width=\linewidth]{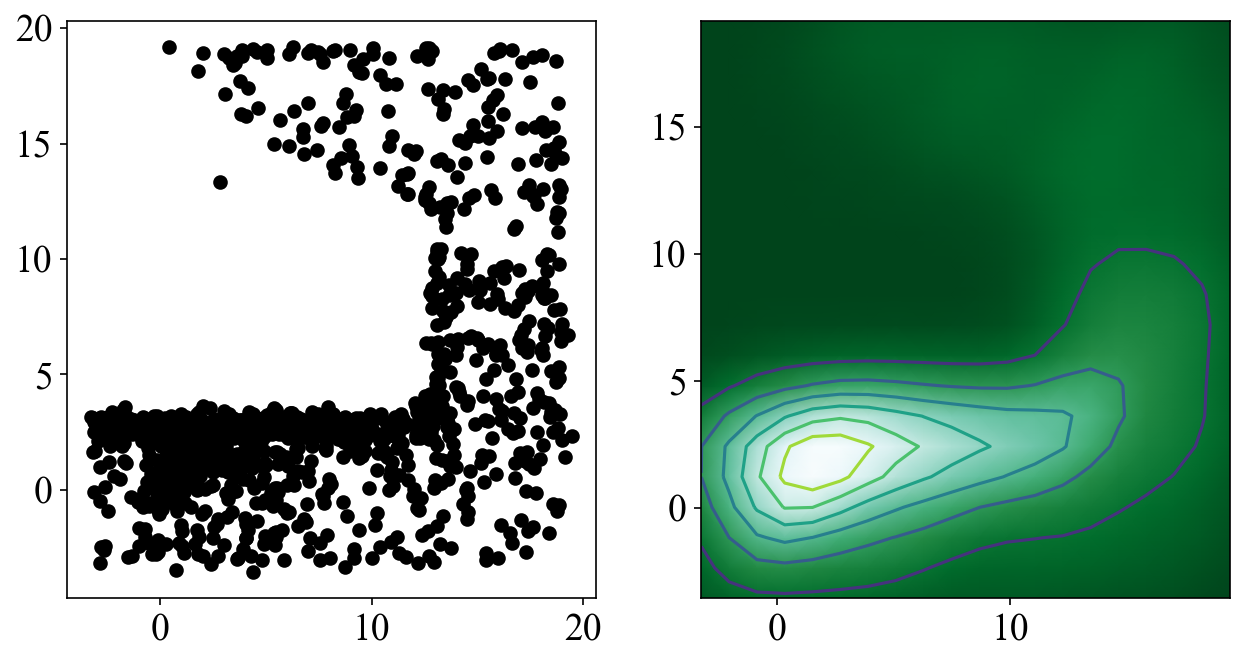}
        \caption{Antmaze HIRO}\label{fig:antmaze_hiro_project}
    \end{subfigure}
    \begin{subfigure}[b]{.45\linewidth}
        \includegraphics[width=\linewidth]{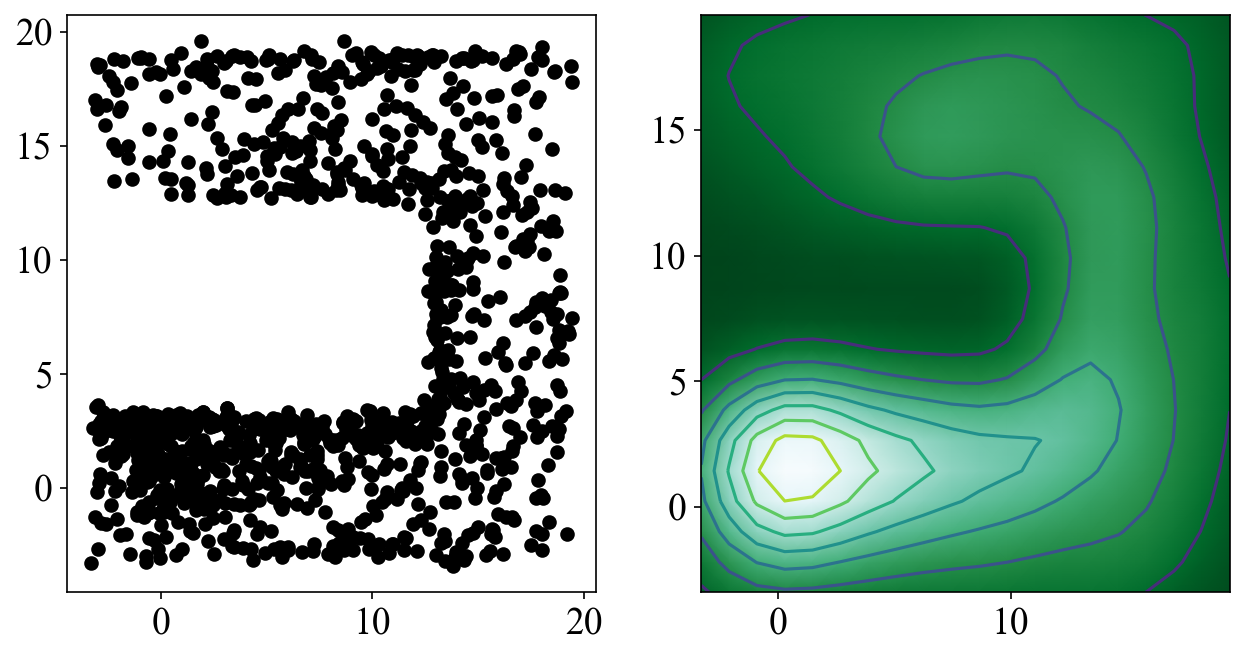}
        \caption{Antmaze Learned Entropy SHIRO}\label{fig:antmaze_shiro_project}
    \end{subfigure}

    \begin{subfigure}[b]{.45\linewidth}
        \includegraphics[width=\linewidth]{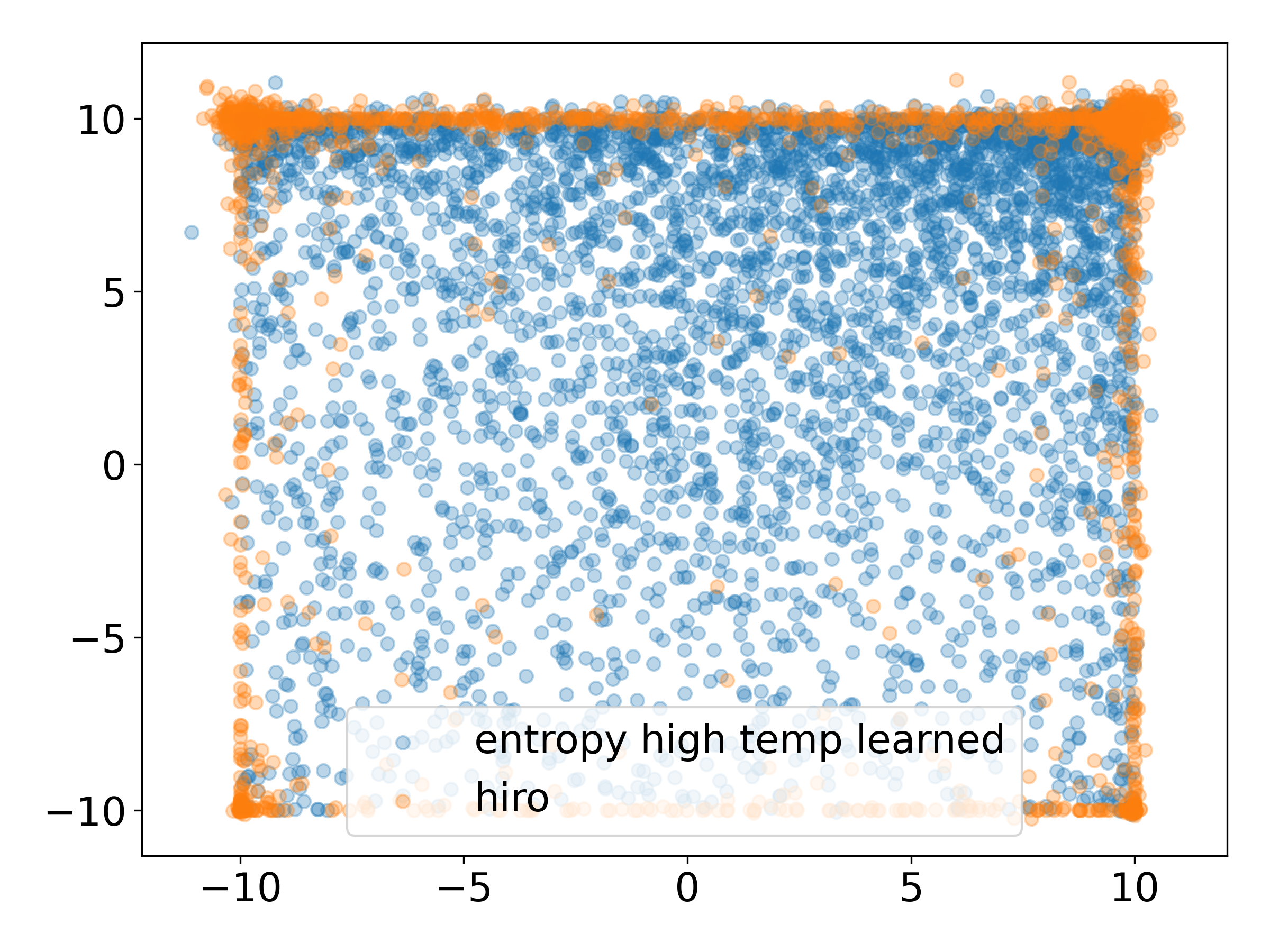}
        \caption{Sub-goal Visualization}\label{fig:sub-goal_viz}
    \end{subfigure}


    \caption{Top: Analysis of the final positions achieved by an agent in a given environment. The left plot is simply a scatterplot of the final positions of an agent on the environment specifies underneath. The second plot displays a contour map showing where the agent ends up on average across the entire experiment. We find that across all environments tested on, SHIRO HL-Learned explores the environment more thoroughly than HIRO. Bottom: X-Y Sub-goal visualization of the high-level policy.}
    \label{fig:project}
\end{figure}

In our research, we used the same environment that was used in the original paper of HIRO \cite{nachum:2018:hiro} to compare against their algorithm. We tested our algorithm on three of their environments: AntMaze, AntPush, and AntFall, as shown in \Cref{fig:env}. All three environments require robotic locomotion and are standard benchmarks of HRL. As presented in \Cref{fig:main_result}, in all environments, our method starts succeeding two to four times faster than the standard HIRO. Although the success rate increases steadily for AntMaze, it fluctuates for AntPush and AntFall, but gradually increases after the drop. This is thought of as a result of exploration for a better policy: Entropy maximization fosters exploration in new regions. The effect of broad exploration is shown in \Cref{fig:antmaze_hiro_project,fig:antmaze_shiro_project}. The figure represents the scatter and contour plot of the final positions achieved by an agent in each environment. From the figure, we can observe that HIRO agent is failing to explore the areas around the goal but repetitively searches around the starting position before eventually learning locomotion to achieve the goal. In comparison to HIRO, qualitatively, our method is being able to reach goals more broadly.

\begin{figure*}[t]
    \centering
    \begin{subfigure}[b]{0.3\textwidth}
        \includegraphics[width=\textwidth]{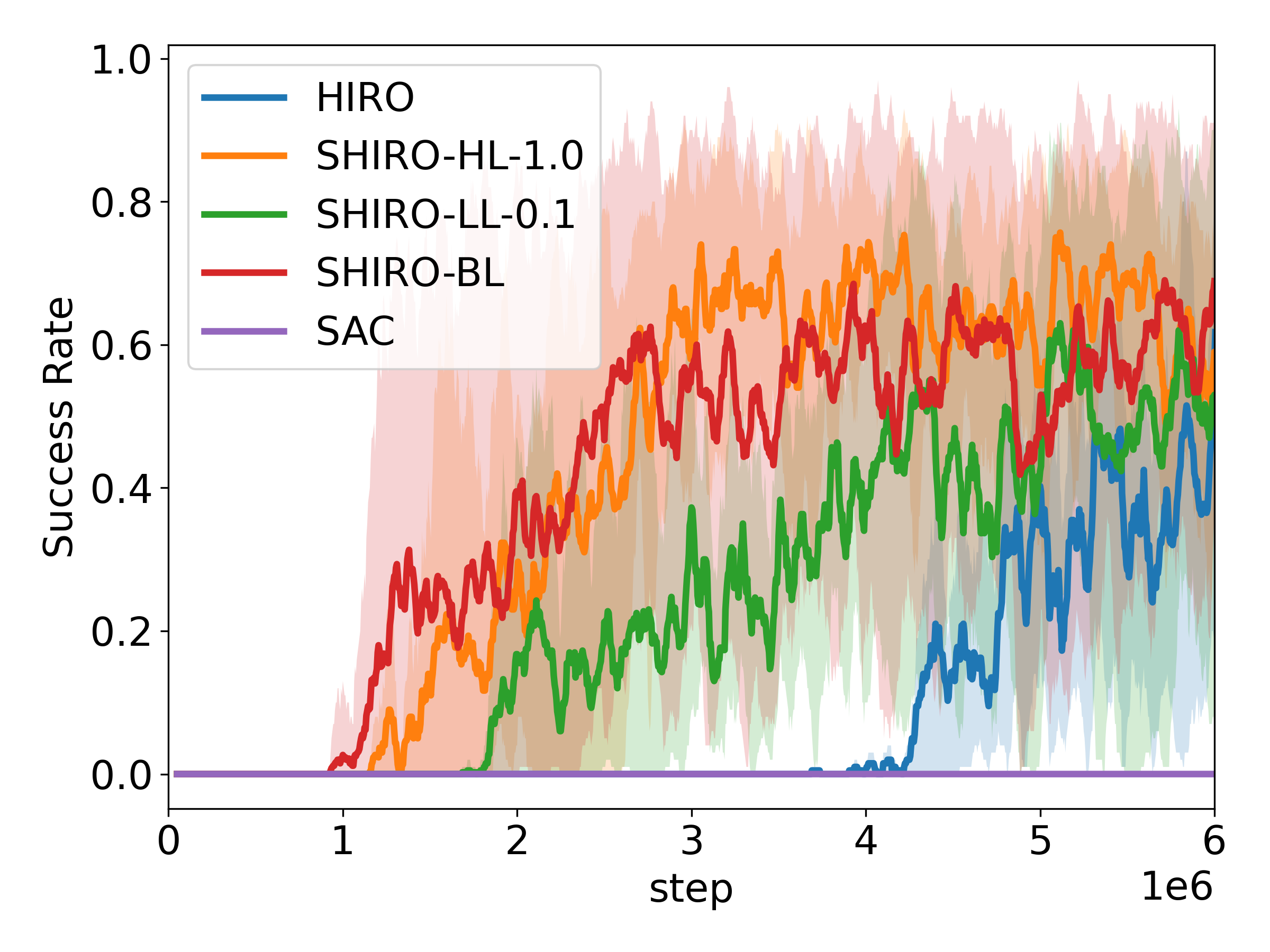}
        \caption{HIRO vs: SHIRO-HL, SHIRO-BL, SHIRO-LL and SAC.}
        \label{fig:ablativestudy_a}
    \end{subfigure}
    \hfill
    \begin{subfigure}[b]{0.3\textwidth}
        \includegraphics[width=\textwidth]{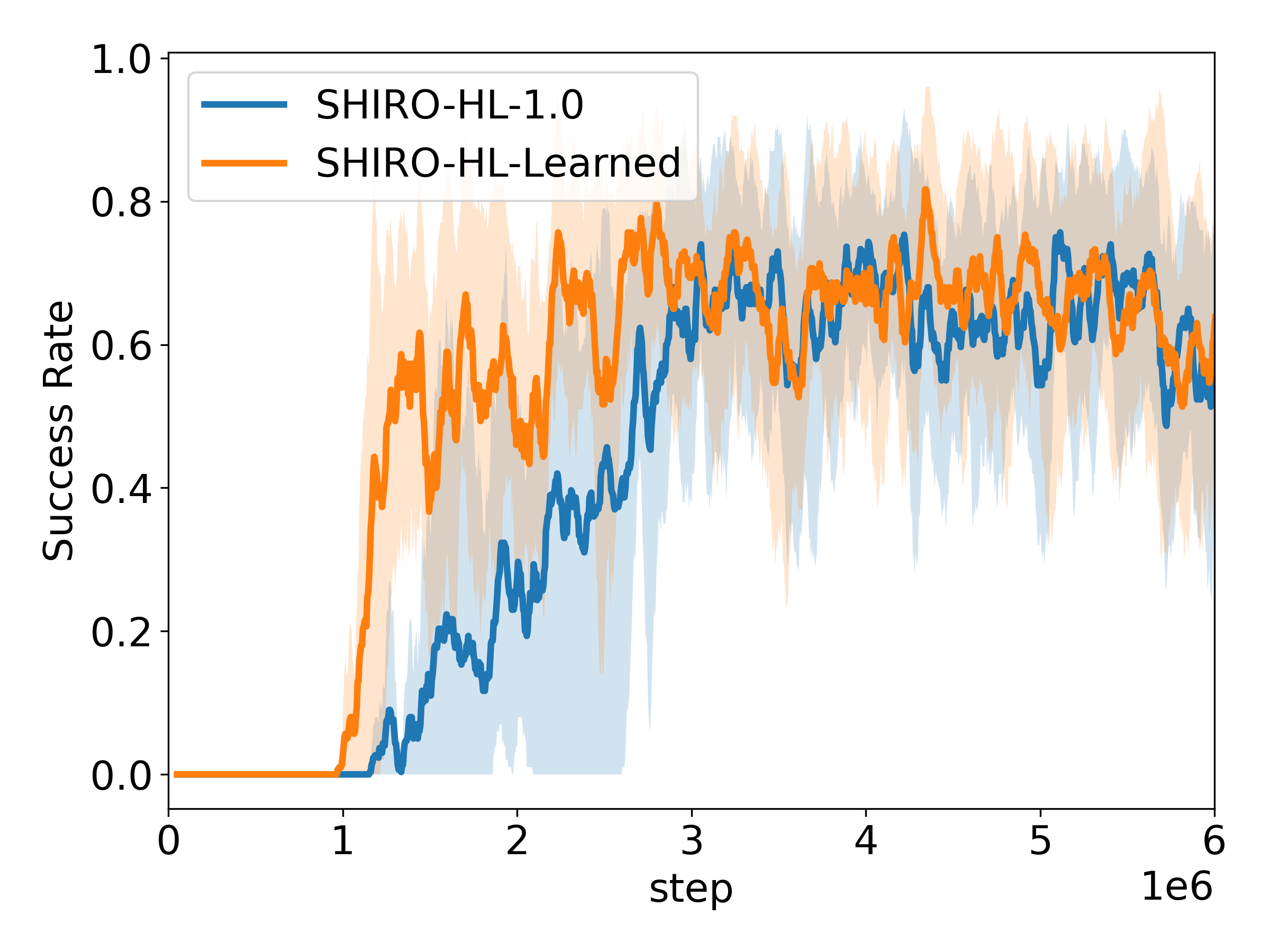}
        \caption{SHIRO HL Learned vs SHIRO HL Const.}
        \label{fig:ablativestudy_b}
    \end{subfigure}
    \hfill
    \begin{subfigure}[b]{0.3\textwidth}
        \includegraphics[width=\textwidth]{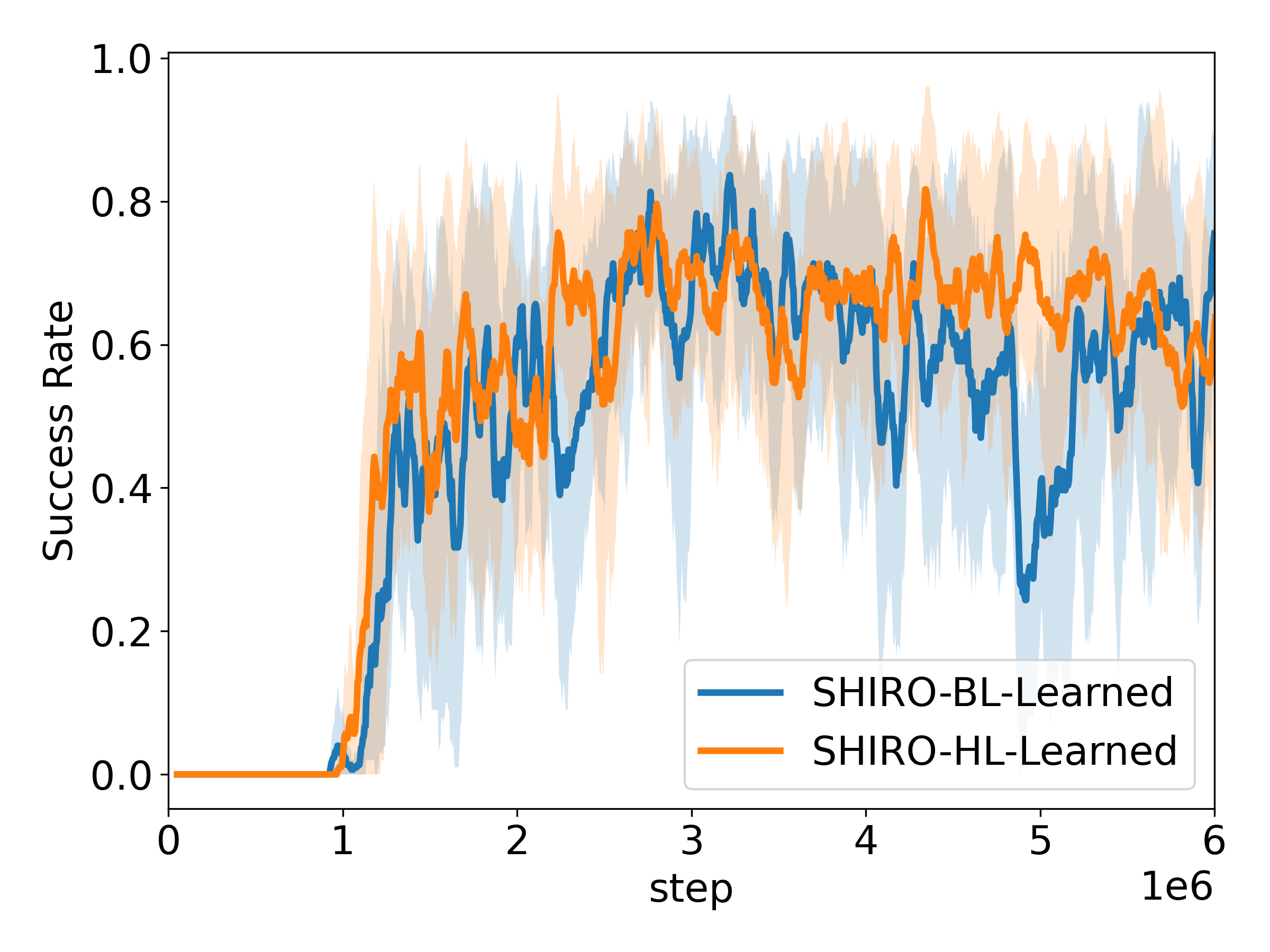}
        \caption{The final duel: SHIRO HL Learned vs SHIRO BL Learned.}
        \label{fig:ablativestudy_c}
    \end{subfigure}
    \caption{Ablative Study conducted in AntMaze. The center lines represent the mean and the shades represent the minimum and maximum range.}
    \label{fig:ablativestudy}
\end{figure*}

\subsection{Ablative Study} \label{sec:ablative}
In \Cref{fig:ablativestudy}, we display the results of our method SHIRO with different configurations that we tested. There are a few different ways to add entropy to a hierarchical agent, so we performed an ablative study to find out the best configuration.

\textbf{Adding Entropy to Each Layer.} We compare agents with entropy on different levels (SHIRO HL/LL/BL), as well as SAC. Each run result consists of three experiments using the same randomly selected seeds for each ablation. \Cref{fig:ablativestudy_a} shows our results. We found that while all ablations perform better than the standard HIRO (besides SAC, which failed to complete the task), some are better than others. Adding entropy to the high-level controller yielded results that were consistently better across the board. Adding entropy to the low-level controller didn't perform as well, starting to succeed faster than regular HIRO, but took longer to reach the maximum success rate (Details in \Cref{sec:klAndLowLevelTemp}). Finally, adding entropy to both levels performed at around the same level as adding it to just the high-level controller.

We find that the learned temperature parameter plays a role in the performance of our ablations, as in \Cref{fig:ablativestudy_b}. If the temperature is high, the agent is incentivized to select actions (sub-goals in the case of the high-level agent) that provide more entropy. With respect to the high-level, while this is good when the agent is starting to learn, we want the agent to start acting deterministically as its success rate reaches its peak.
We can observe this behavior in one of the figures in \Cref{sec:Additional_Figures}, where the high-level temperature generally decreases with time according to the performance of the agent.

\textbf{High-Level vs. Both-Levels} Since both SHIRO HL-Learned and BL-Learned perform well, it is necessary to compare them in order to find the best option, shown in \Cref{fig:ablativestudy_c}. We find that SHIRO HL-Learned performs marginally better than SHIRO BL-Learned. Thus, we consider SHIRO HL-Learned to be our best method produced from the ablative study, as well as the fact that SHIRO-HL-Learned does not require any temperature parameter tuning on the low-level. However, we suggest that both methods are viable, and one or the other can be chosen according to preference.

\subsection{Analysis} \label{sec:klAndLowLevelTemp}
In our experiments, the KL-divergence in the low-level policy is sufficiently low in our ablations, which follows  \Cref{theorem:KLbound} that can be applied to HIRO with and without high-level entropy. In general, the KL divergence is below $1.0$ (a plot can be found in \Cref{sec:Additional_Figures}). However, even though SHIRO with a low-level temperature of $1.0$ had the lowest KL divergence, it fails to learn because the high temperature parameter induces Q-value overestimation. As the temperature increases, the second term in \Cref{eq:valueFunc} becomes large that overestimates the actions that are unlikely to be taken. It also creates a too stochastic environment for the high-level agent to solve the main task.





We also observe that the addition of entropy to the high-level agent significantly improves the learning of the \textit{low-level} agent, in turn, improving the high-level agent by making the environment more ``helpful'' to solve the main task, as well as being able to broadly explore the environment. The more diverse sub-goals provided to the low-level agent increase the loss (see \Cref{sec:Additional_Figures}) of that agent's Q-functions because the Q-functions must initially learn from more diverse data (this difference in diversity can be seen in \Cref{fig:sub-goal_viz}). This allows for larger updates (from gradient descent) that enable the Q-function to learn from underrepresented goals.
With a more robust Q-function that has received more meaningful learning signals, we also observe an improvement in the low-level policy loss (details in \Cref{sec:Additional_Figures}), where the policy directly optimizes from the Q-function. In turn, the low-level agent learns locomotion quickly.

Another observation from our study is that high-level entropy maximization improves slightly ``worse'' low-level agents. When the high-level agent is kept deterministic, selecting a low-level policy as SAC (which has empirically shown to outperform TD3) demonstrates better performance than a TD3 low-level policy. However, when the high-level policy maximizes entropy, the performance of different low-level agents significantly improves but are similar to one another, suggesting that the diversity of data significantly improves worse low-level agents. Graphs of these observations are located in \Cref{sec:Additional_Figures}.

\section{Conclusion} \label{sec:Conclusion}





In this work, we proposed an effective solution to the problem of redundant sub-goals for robotic learning in goal-conditioned hierarchical reinforcement learning. We achieved this through maximum entropy goal-conditioned HRL, in which we treated it as a single agent when the KL divergence is small between the low-level policy updates. Our empirical findings motivate the learning of entropy temperature, namely on the high-level. We performed an ablative study to understand the effects of entropy on both levels, where high-level learned entropy had significant impact on the efficiency. Concretely, the diversity of sub-goals provided to the low-level agent improved the redundancy and reachability of goals, allowing for a high-level agent to learn even faster. Regarding future work, we are interested in applying our findings on a real robot platform, as well as exploration of the importance of sub-goal representations.

\bibliography{main}

\begin{thebibliography}{61}
\providecommand{\natexlab}[1]{#1}
\providecommand{\url}[1]{\texttt{#1}}
\expandafter\ifx\csname urlstyle\endcsname\relax
  \providecommand{\doi}[1]{doi: #1}\else
  \providecommand{\doi}{doi: \begingroup \urlstyle{rm}\Url}\fi

\bibitem[Gu et~al.(2017)Gu, Holly, Lillicrap, and
  Levine]{gu:2017:DRL4Manipulation}
S.~Gu, E.~Holly, T.~Lillicrap, and S.~Levine.
\newblock Deep reinforcement learning for robotic manipulation with
  asynchronous off-policy updates.
\newblock In \emph{2017 IEEE international conference on robotics and
  automation (ICRA)}, pages 3389--3396. IEEE, 2017.

\bibitem[Yang et~al.(2020)Yang, Caluwaerts, Iscen, Zhang, Tan, and
  Sindhwani]{yang:2020:HIRO4Legged}
Y.~Yang, K.~Caluwaerts, A.~Iscen, T.~Zhang, J.~Tan, and V.~Sindhwani.
\newblock Data efficient reinforcement learning for legged robots.
\newblock In \emph{Conference on Robot Learning}, pages 1--10. PMLR, 2020.

\bibitem[Jain et~al.(2019)Jain, Iscen, and Caluwaerts]{jain:2019:HRL4Legged}
D.~Jain, A.~Iscen, and K.~Caluwaerts.
\newblock Hierarchical reinforcement learning for quadruped locomotion.
\newblock In \emph{2019 IEEE/RSJ International Conference on Intelligent Robots
  and Systems (IROS)}, pages 7551--7557. IEEE, 2019.

\bibitem[Duan et~al.(2016)Duan, Chen, Houthooft, Schulman, and
  Abbeel]{duan:2016:benchmarking}
Y.~Duan, X.~Chen, R.~Houthooft, J.~Schulman, and P.~Abbeel.
\newblock Benchmarking deep reinforcement learning for continuous control.
\newblock In \emph{International conference on machine learning}, pages
  1329--1338. PMLR, 2016.

\bibitem[Peng et~al.(2018)Peng, Abbeel, Levine, and van~de
  Panne]{peng:2018:deepmimic}
X.~B. Peng, P.~Abbeel, S.~Levine, and M.~van~de Panne.
\newblock Deepmimic: Example-guided deep reinforcement learning of
  physics-based character skills.
\newblock \emph{ACM Transactions on Graphics (TOG)}, 37\penalty0 (4):\penalty0
  1--14, 2018.

\bibitem[Peng et~al.(2019)Peng, Chang, Zhang, Abbeel, and
  Levine]{peng:2019:mcp}
X.~B. Peng, M.~Chang, G.~Zhang, P.~Abbeel, and S.~Levine.
\newblock Mcp: learning composable hierarchical control with multiplicative
  compositional policies.
\newblock In \emph{Proceedings of the 33rd International Conference on Neural
  Information Processing Systems}, pages 3686--3697, 2019.

\bibitem[Silver et~al.(2016)Silver, Huang, Maddison, Guez, Sifre, Van
  Den~Driessche, Schrittwieser, Antonoglou, Panneershelvam, Lanctot,
  et~al.]{silver:2016:alphago}
D.~Silver, A.~Huang, C.~J. Maddison, A.~Guez, L.~Sifre, G.~Van Den~Driessche,
  J.~Schrittwieser, I.~Antonoglou, V.~Panneershelvam, M.~Lanctot, et~al.
\newblock Mastering the game of go with deep neural networks and tree search.
\newblock \emph{nature}, 529\penalty0 (7587):\penalty0 484--489, 2016.

\bibitem[Nachum et~al.(2018)Nachum, Gu, Lee, and Levine]{nachum:2018:hiro}
O.~Nachum, S.~S. Gu, H.~Lee, and S.~Levine.
\newblock Data-efficient hierarchical reinforcement learning.
\newblock \emph{Advances in Neural Information Processing Systems},
  31:\penalty0 3303--3313, 2018.

\bibitem[Levy et~al.(2018)Levy, Konidaris, Platt, and
  Saenko]{levy:2018:multi-level-hindsight}
A.~Levy, G.~Konidaris, R.~Platt, and K.~Saenko.
\newblock Learning multi-level hierarchies with hindsight.
\newblock In \emph{International Conference on Learning Representations}, 2018.

\bibitem[Jiang et~al.(2019)Jiang, Gu, Murphy, and
  Finn]{jiang:2019:language-as-abstraction}
Y.~Jiang, S.~S. Gu, K.~P. Murphy, and C.~Finn.
\newblock Language as an abstraction for hierarchical deep reinforcement
  learning.
\newblock \emph{Advances in Neural Information Processing Systems},
  32:\penalty0 9419--9431, 2019.

\bibitem[Nachum et~al.(2018)Nachum, Gu, Lee, and
  Levine]{nachum:2018:nearoptimal}
O.~Nachum, S.~Gu, H.~Lee, and S.~Levine.
\newblock Near-optimal representation learning for hierarchical reinforcement
  learning.
\newblock In \emph{International Conference on Learning Representations}, 2018.

\bibitem[Dayan and Hinton(1993)]{dayan:1993:feudal}
P.~Dayan and G.~E. Hinton.
\newblock Feudal reinforcement learning.
\newblock In S.~Hanson, J.~Cowan, and C.~Giles, editors, \emph{Advances in
  Neural Information Processing Systems}, volume~5. Morgan-Kaufmann, 1993.

\bibitem[Parr and Russell(1998)]{parr:1998:hierarchiesofmachines}
R.~Parr and S.~Russell.
\newblock Reinforcement learning with hierarchies of machines.
\newblock \emph{Advances in neural information processing systems}, pages
  1043--1049, 1998.

\bibitem[Sutton et~al.(1999)Sutton, Precup, and Singh]{sutton:1999:betweenmdp}
R.~S. Sutton, D.~Precup, and S.~Singh.
\newblock Between mdps and semi-mdps: A framework for temporal abstraction in
  reinforcement learning.
\newblock \emph{Artificial intelligence}, 112\penalty0 (1-2):\penalty0
  181--211, 1999.

\bibitem[Dietterich(2000)]{dietterich:2000:MAXQ}
T.~G. Dietterich.
\newblock Hierarchical reinforcement learning with the maxq value function
  decomposition.
\newblock \emph{Journal of artificial intelligence research}, 13:\penalty0
  227--303, 2000.

\bibitem[Nachum et~al.(2019)Nachum, Tang, Lu, Gu, Lee, and
  Levine]{nachum2019does}
O.~Nachum, H.~Tang, X.~Lu, S.~Gu, H.~Lee, and S.~Levine.
\newblock Why does hierarchy (sometimes) work so well in reinforcement
  learning?
\newblock \emph{arXiv preprint arXiv:1909.10618}, 2019.

\bibitem[Zhang et~al.(2020)Zhang, Guo, Tan, Hu, and Chen]{zhang2020generating}
T.~Zhang, S.~Guo, T.~Tan, X.~Hu, and F.~Chen.
\newblock Generating adjacency-constrained subgoals in hierarchical
  reinforcement learning.
\newblock \emph{Advances in Neural Information Processing Systems}, 33, 2020.

\bibitem[Li et~al.(2021)Li, Zheng, Wang, and Zhang]{lilearning}
S.~Li, L.~Zheng, J.~Wang, and C.~Zhang.
\newblock Learning subgoal representations with slow dynamics.
\newblock In \emph{International Conference on Learning Representations}, 2021.

\bibitem[Haarnoja et~al.(2018)Haarnoja, Zhou, Abbeel, and
  Levine]{haarnoja:2018:sac}
T.~Haarnoja, A.~Zhou, P.~Abbeel, and S.~Levine.
\newblock Soft actor-critic: Off-policy maximum entropy deep reinforcement
  learning with a stochastic actor.
\newblock In \emph{International Conference on Machine Learning}, pages
  1861--1870. PMLR, 2018.

\bibitem[Ziebart et~al.(2008)Ziebart, Maas, Bagnell, and
  Dey]{ziebart2008maximum}
B.~D. Ziebart, A.~L. Maas, J.~A. Bagnell, and A.~K. Dey.
\newblock Maximum entropy inverse reinforcement learning.
\newblock In \emph{Aaai}, volume~8, pages 1433--1438. Chicago, IL, USA, 2008.

\bibitem[Toussaint(2009)]{toussaint2009robot}
M.~Toussaint.
\newblock Robot trajectory optimization using approximate inference.
\newblock In \emph{Proceedings of the 26th annual international conference on
  machine learning}, pages 1049--1056, 2009.

\bibitem[Rawlik et~al.(2012)Rawlik, Toussaint, and
  Vijayakumar]{rawlik2012stochastic}
K.~Rawlik, M.~Toussaint, and S.~Vijayakumar.
\newblock On stochastic optimal control and reinforcement learning by
  approximate inference.
\newblock \emph{Proceedings of Robotics: Science and Systems VIII}, 2012.

\bibitem[Fox et~al.(2016)Fox, Pakman, and Tishby]{fox2016taming}
R.~Fox, A.~Pakman, and N.~Tishby.
\newblock Taming the noise in reinforcement learning via soft updates.
\newblock In \emph{Proceedings of the Thirty-Second Conference on Uncertainty
  in Artificial Intelligence}, pages 202--211, 2016.

\bibitem[Haarnoja et~al.(2017)Haarnoja, Tang, Abbeel, and
  Levine]{haarnoja2017reinforcement}
T.~Haarnoja, H.~Tang, P.~Abbeel, and S.~Levine.
\newblock Reinforcement learning with deep energy-based policies.
\newblock In \emph{International Conference on Machine Learning}, pages
  1352--1361. PMLR, 2017.

\bibitem[Levine(2018)]{levine2018reinforcement}
S.~Levine.
\newblock Reinforcement learning and control as probabilistic inference:
  Tutorial and review.
\newblock \emph{arXiv preprint arXiv:1805.00909}, 2018.

\bibitem[Stolle and Precup(2002)]{stolle:2002:learningoptions}
M.~Stolle and D.~Precup.
\newblock Learning options in reinforcement learning.
\newblock In \emph{International Symposium on abstraction, reformulation, and
  approximation}, pages 212--223. Springer, 2002.

\bibitem[Mannor et~al.(2004)Mannor, Menache, Hoze, and
  Klein]{mannor:2004:dynamic}
S.~Mannor, I.~Menache, A.~Hoze, and U.~Klein.
\newblock Dynamic abstraction in reinforcement learning via clustering.
\newblock In \emph{Proceedings of the twenty-first international conference on
  Machine learning}, page~71, 2004.

\bibitem[Singh et~al.(2005)Singh, Barto, and
  Chentanez]{singh:2005:intrinsically}
S.~Singh, A.~G. Barto, and N.~Chentanez.
\newblock Intrinsically motivated reinforcement learning.
\newblock Technical report, MASSACHUSETTS UNIV AMHERST DEPT OF COMPUTER
  SCIENCE, 2005.

\bibitem[Precup(2000)]{precup:2000:temporal}
D.~Precup.
\newblock \emph{Temporal abstraction in reinforcement learning}.
\newblock University of Massachusetts Amherst, 2000.

\bibitem[Bacon et~al.(2017)Bacon, Harb, and Precup]{bacon:2017:option}
P.-L. Bacon, J.~Harb, and D.~Precup.
\newblock The option-critic architecture.
\newblock In \emph{Proceedings of the Thirty-First AAAI Conference on
  Artificial Intelligence}, AAAI'17, page 1726–1734. AAAI Press, 2017.

\bibitem[Harb et~al.(2018)Harb, Bacon, Klissarov, and
  Precup]{harb:2018:waiting}
J.~Harb, P.-L. Bacon, M.~Klissarov, and D.~Precup.
\newblock When waiting is not an option: Learning options with a deliberation
  cost.
\newblock In \emph{Proceedings of the AAAI Conference on Artificial
  Intelligence}, volume~32, 2018.

\bibitem[Vezhnevets et~al.(2016)Vezhnevets, Mnih, Osindero, Graves, Vinyals,
  Agapiou, et~al.]{mnih:2016:strategic}
A.~Vezhnevets, V.~Mnih, S.~Osindero, A.~Graves, O.~Vinyals, J.~Agapiou, et~al.
\newblock Strategic attentive writer for learning macro-actions.
\newblock \emph{Advances in Neural Information Processing Systems},
  29:\penalty0 3486--3494, 2016.

\bibitem[Frans et~al.(2018)Frans, Ho, Chen, Abbeel, and
  Schulman]{frans:2018:meta}
K.~Frans, J.~Ho, X.~Chen, P.~Abbeel, and J.~Schulman.
\newblock Meta learning shared hierarchies.
\newblock In \emph{International Conference on Learning Representations}, 2018.

\bibitem[Sigaud and Stulp(2019)]{sigaud:2019:policy}
O.~Sigaud and F.~Stulp.
\newblock Policy search in continuous action domains: an overview.
\newblock \emph{Neural Networks}, 113:\penalty0 28--40, 2019.

\bibitem[Heess et~al.(2016)Heess, Wayne, Tassa, Lillicrap, Riedmiller, and
  Silver]{heess:2016:learning}
N.~Heess, G.~Wayne, Y.~Tassa, T.~Lillicrap, M.~Riedmiller, and D.~Silver.
\newblock Learning and transfer of modulated locomotor controllers.
\newblock \emph{arXiv preprint arXiv:1610.05182}, 2016.

\bibitem[Kulkarni et~al.(2016)Kulkarni, Narasimhan, Saeedi, and
  Tenenbaum]{kulkarni:2016:hierarchical}
T.~D. Kulkarni, K.~Narasimhan, A.~Saeedi, and J.~Tenenbaum.
\newblock Hierarchical deep reinforcement learning: Integrating temporal
  abstraction and intrinsic motivation.
\newblock \emph{Advances in neural information processing systems},
  29:\penalty0 3675--3683, 2016.

\bibitem[Tessler et~al.(2017)Tessler, Givony, Zahavy, Mankowitz, and
  Mannor]{tessler:2017:deep}
C.~Tessler, S.~Givony, T.~Zahavy, D.~J. Mankowitz, and S.~Mannor.
\newblock A deep hierarchical approach to lifelong learning in minecraft.
\newblock In \emph{Proceedings of the Thirty-First AAAI Conference on
  Artificial Intelligence}, AAAI'17, page 1553–1561. AAAI Press, 2017.

\bibitem[Florensa et~al.(2017)Florensa, Duan, and
  Abbeel]{florensa:2017:stochastic}
C.~Florensa, Y.~Duan, and P.~Abbeel.
\newblock Stochastic neural networks for hierarchical reinforcement learning.
\newblock In \emph{ICLR (Poster)}, 2017.

\bibitem[Konidaris and Barto(2007)]{konidaris:2007:building}
G.~D. Konidaris and A.~G. Barto.
\newblock Building portable options: Skill transfer in reinforcement learning.
\newblock In \emph{IJCAI}, volume~7, pages 895--900, 2007.

\bibitem[Daniel et~al.(2012)Daniel, Neumann, and
  Peters]{daniel:2012:hierarchical}
C.~Daniel, G.~Neumann, and J.~Peters.
\newblock Hierarchical relative entropy policy search.
\newblock In \emph{Artificial Intelligence and Statistics}, pages 273--281.
  PMLR, 2012.

\bibitem[Schaul et~al.(2015)Schaul, Horgan, Gregor, and
  Silver]{schaul:2015:universal}
T.~Schaul, D.~Horgan, K.~Gregor, and D.~Silver.
\newblock Universal value function approximators.
\newblock In \emph{International conference on machine learning}, pages
  1312--1320. PMLR, 2015.

\bibitem[Andrychowicz et~al.(2017)Andrychowicz, Wolski, Ray, Schneider, Fong,
  Welinder, McGrew, Tobin, Abbeel, and Zaremba]{andrychowicz:2017:hindsight}
M.~Andrychowicz, F.~Wolski, A.~Ray, J.~Schneider, R.~Fong, P.~Welinder,
  B.~McGrew, J.~Tobin, P.~Abbeel, and W.~Zaremba.
\newblock Hindsight experience replay.
\newblock In \emph{Proceedings of the 31st International Conference on Neural
  Information Processing Systems}, pages 5055--5065, 2017.

\bibitem[Vezhnevets et~al.(2017)Vezhnevets, Osindero, Schaul, Heess, Jaderberg,
  Silver, and Kavukcuoglu]{vezhnevets:2017:feudal}
A.~S. Vezhnevets, S.~Osindero, T.~Schaul, N.~Heess, M.~Jaderberg, D.~Silver,
  and K.~Kavukcuoglu.
\newblock Feudal networks for hierarchical reinforcement learning.
\newblock In \emph{International Conference on Machine Learning}, pages
  3540--3549. PMLR, 2017.

\bibitem[Nachum et~al.(2020)Nachum, Ahn, Ponte, Gu, and Kumar]{nachum2020multi}
O.~Nachum, M.~Ahn, H.~Ponte, S.~S. Gu, and V.~Kumar.
\newblock Multi-agent manipulation via locomotion using hierarchical sim2real.
\newblock In \emph{Conference on Robot Learning}, pages 110--121. PMLR, 2020.

\bibitem[Mahadevan and Maggioni(2007)]{mahadevan:2007:proto}
S.~Mahadevan and M.~Maggioni.
\newblock Proto-value functions: A laplacian framework for learning
  representation and control in markov decision processes.
\newblock \emph{Journal of Machine Learning Research}, 8\penalty0 (10), 2007.

\bibitem[Sutton et~al.(2011)Sutton, Modayil, Delp, Degris, Pilarski, White, and
  Precup]{sutton:2011:horde}
R.~S. Sutton, J.~Modayil, M.~Delp, T.~Degris, P.~M. Pilarski, A.~White, and
  D.~Precup.
\newblock Horde: A scalable real-time architecture for learning knowledge from
  unsupervised sensorimotor interaction.
\newblock In \emph{The 10th International Conference on Autonomous Agents and
  Multiagent Systems-Volume 2}, pages 761--768, 2011.

\bibitem[Pong et~al.(2018)Pong, Gu, Dalal, and Levine]{pong:2018:temporal}
V.~Pong, S.~Gu, M.~Dalal, and S.~Levine.
\newblock Temporal difference models: Model-free deep rl for model-based
  control.
\newblock In \emph{International Conference on Learning Representations}, 2018.

\bibitem[Li et~al.(2019)Li, Florensa, Clavera, and Abbeel]{li2019sub}
A.~Li, C.~Florensa, I.~Clavera, and P.~Abbeel.
\newblock Sub-policy adaptation for hierarchical reinforcement learning.
\newblock In \emph{International Conference on Learning Representations}, 2019.

\bibitem[Hejna et~al.(2020)Hejna, Pinto, and Abbeel]{hejna2020hierarchically}
D.~Hejna, L.~Pinto, and P.~Abbeel.
\newblock Hierarchically decoupled imitation for morphological transfer.
\newblock In \emph{International Conference on Machine Learning}, pages
  4159--4171. PMLR, 2020.

\bibitem[Zhang et~al.(2020)Zhang, Yu, and Xu]{zhang2020hierarchical}
J.~Zhang, H.~Yu, and W.~Xu.
\newblock Hierarchical reinforcement learning by discovering intrinsic options.
\newblock In \emph{International Conference on Learning Representations}, 2020.

\bibitem[Li et~al.(2019)Li, Wang, Tang, and Zhang]{li2019hierarchical}
S.~Li, R.~Wang, M.~Tang, and C.~Zhang.
\newblock Hierarchical reinforcement learning with advantage-based auxiliary
  rewards.
\newblock In \emph{Proceedings of the 33rd International Conference on Neural
  Information Processing Systems}, pages 1409--1419, 2019.

\bibitem[Zhao et~al.(2019)Zhao, Sun, and Tresp]{zhao2019maximum}
R.~Zhao, X.~Sun, and V.~Tresp.
\newblock Maximum entropy-regularized multi-goal reinforcement learning.
\newblock In \emph{International Conference on Machine Learning}, pages
  7553--7562. PMLR, 2019.

\bibitem[Haarnoja et~al.(2018)Haarnoja, Hartikainen, Abbeel, and
  Levine]{haarnoja2018latent}
T.~Haarnoja, K.~Hartikainen, P.~Abbeel, and S.~Levine.
\newblock Latent space policies for hierarchical reinforcement learning.
\newblock In \emph{International Conference on Machine Learning}, pages
  1851--1860. PMLR, 2018.

\bibitem[Azarafrooz and Brock(2019)]{azarafrooz2019hierarchical}
A.~Azarafrooz and J.~Brock.
\newblock Hierarchical soft actor-critic: Adversarial exploration via mutual
  information optimization.
\newblock \emph{arXiv preprint arXiv:1906.07122}, 2019.

\bibitem[Tang et~al.(2021)Tang, Wang, Xue, Yang, and Cao]{tang2021novel}
H.~Tang, A.~Wang, F.~Xue, J.~Yang, and Y.~Cao.
\newblock A novel hierarchical soft actor-critic algorithm for multi-logistics
  robots task allocation.
\newblock \emph{IEEE Access}, 9:\penalty0 42568--42582, 2021.

\bibitem[Beyret et~al.(2019)Beyret, Shafti, and Faisal]{beyret:2019:dot2dot}
B.~Beyret, A.~Shafti, and A.~A. Faisal.
\newblock Dot-to-dot: Explainable hierarchical reinforcement learning for
  robotic manipulation.
\newblock In \emph{2019 IEEE/RSJ International Conference on Intelligent Robots
  and Systems (IROS)}, pages 5014--5019. IEEE, 2019.

\bibitem[Lillicrap et~al.(2016)Lillicrap, Hunt, Pritzel, Heess, Erez, Tassa,
  Silver, and Wierstra]{lillicrap2016continuous}
T.~P. Lillicrap, J.~J. Hunt, A.~Pritzel, N.~Heess, T.~Erez, Y.~Tassa,
  D.~Silver, and D.~Wierstra.
\newblock Continuous control with deep reinforcement learning.
\newblock In \emph{ICLR (Poster)}, 2016.

\bibitem[Fujimoto et~al.(2018)Fujimoto, Hoof, and
  Meger]{fujimoto2018addressing}
S.~Fujimoto, H.~Hoof, and D.~Meger.
\newblock Addressing function approximation error in actor-critic methods.
\newblock In \emph{International Conference on Machine Learning}, pages
  1587--1596. PMLR, 2018.

\bibitem[Haarnoja et~al.(2018)Haarnoja, Zhou, Hartikainen, Tucker, Ha, Tan,
  Kumar, Zhu, Gupta, Abbeel, et~al.]{haarnoja2018soft}
T.~Haarnoja, A.~Zhou, K.~Hartikainen, G.~Tucker, S.~Ha, J.~Tan, V.~Kumar,
  H.~Zhu, A.~Gupta, P.~Abbeel, et~al.
\newblock Soft actor-critic algorithms and applications.
\newblock \emph{arXiv preprint arXiv:1812.05905}, 2018.

\bibitem[Fujita et~al.(2021)Fujita, Nagarajan, Kataoka, and
  Ishikawa]{JMLR:v22:20-376}
Y.~Fujita, P.~Nagarajan, T.~Kataoka, and T.~Ishikawa.
\newblock Chainerrl: A deep reinforcement learning library.
\newblock \emph{Journal of Machine Learning Research}, 22\penalty0
  (77):\penalty0 1--14, 2021.
\newblock URL \url{http://jmlr.org/papers/v22/20-376.html}.

\bibitem[Abdolmaleki et~al.(2018)Abdolmaleki, Springenberg, Tassa, Munos,
  Heess, and Riedmiller]{abdolmaleki2018maximum}
A.~Abdolmaleki, J.~T. Springenberg, Y.~Tassa, R.~Munos, N.~Heess, and
  M.~Riedmiller.
\newblock Maximum a posteriori policy optimisation.
\newblock In \emph{International Conference on Learning Representations}, 2018.

\end{thebibliography}

\newpage 

\appendix
\section{Appendix A: Proofs and Additional Derivations} \label{sec:Proofs}

\subsection{Proof of \texorpdfstring{\Cref{theorem:KLbound}}{KL bounding proof}}

\begin{proof}
In this proof, we show that the change in the Abstract Transition function $P_{T \pi^l_{\phi}}^{abs}(s_{t+c} | s_t, g_t)$ can be bounded if the policy is updated such that the policies before and after the parameter update after time step $c$ are \emph{close}. 
In the deterministic case, we say that the policy $\pi^l_{\phi'}$ after the update is \textit{close} to the policy $\pi^l_{\phi}$ before the update if taking different action from the old policy is bounded by a small probability value  $\pi^l_{\phi'}(a_t \neq \pi^l_{\phi}(s_t, g_t) | s_t, g_t) \leq \epsilon$.
Let the probability of arriving at state $s_{t+c}$ starting from state $s_t$ given sub-goal $g_t$ be

\begin{equation}
    p_{\phi'}(s_{t+c} | s_t, g_t) = (1-\epsilon)^c p_{\phi} (s_{t+c}|s_t, g_t) + (1- (1-\epsilon)^c) p_{\text{mistakes}} (s_{t+c} | s_t, g_t)
\end{equation}

The $p_{\phi'}(s_{t+c} | s_t, g)$ is defined as an addition of the probability of not making mistake for $c$ consecutive steps and the probability of making at least one mistake, where $p_{\text{mistakes}}$ is some distribution that makes mistakes.
Then the difference in the probability, defined as total variation divergence between $p_{\phi'}$ and $p_{\phi}$, is bounded by $2\epsilon c$.

\begin{align}
    |p_{\phi'}(s_{t+c} | s_t, g_t) - p_{\phi} (s_{t+c}|s_t, g_t)| &= (1- (1-\epsilon)^c) |p_{\text{mistakes}} (s_{t+c} | s_t, g_t) - p_{\phi} (s_{t+c}|s_t, g_t) | \\
    &\leq 2(1- (1-\epsilon)^c) \\
    &\leq 2\epsilon c
\end{align}

Thus, as long as the probability of a policy making a mistake is small, then the probability of transitioning to $s_{t+c}$ is almost the same as before. Notice that $ p_{\phi} (s_{t+c}|s_t, g_t)$ is exactly equal to $P_{T \pi^l_{\phi}}^{abs}(s_{t+c} | s_t, g_t)$. Thus, we can say that the change in the Abstracted Transition function does not change much if policy stays similar. In fact, we can similarly derive a bound for the stochastic case as in \cite{levine2018reinforcement}. As long as the total variation divergence between the two policies is bounded by $\epsilon$, i.e., $| \pi^l_{\phi'} (a_t | s_t, g_t) - \pi^l_{\phi} (a_t | s_t, g_t) | \leq \epsilon$ for all $s_t$, then the policies are said to be \textit{close}.
\end{proof}


\subsection{KL Divergence of the Deterministic Policy with Gaussian Noise}

In our work, we computed the KL divergence between low-level policies to compare the results against other methods and to verify our theoretical findings.
For deterministic policies, especially policies with Gaussian exploration noise, we view them as stochastic and derive the KL divergence as follows. Let $\mu: \mathcal{S} \times \mathcal{G} \rightarrow \mathcal{A}$ be the deterministic head of the policy and $\Sigma$ be the exploration noise.

\begin{equation}
    \pi(a_t | s_t, g_t) = \mathcal{N}(\mu(s_t, g_t), \Sigma),
\end{equation}

Thus, under this perspective, the KL divergence between a old policy $\pi_{\phi}$ and new $\pi_{\phi'}$ after a parameter update can be expressed as:

\begin{equation}
    D_{KL}(\pi_{\phi}||\pi_{\phi'}) = \dfrac{1}{2}\left[\log{\frac{|\Sigma_{\phi'}|}{|\Sigma_{\phi}|}} - k + (\mu_{\phi'} - \mu_{\phi})^T{\Sigma}_{\phi'}^{-1}(\mu_{\phi'} - \mu_{\phi}) + tr\{{{\Sigma}_{\phi'}^{-1}}{\Sigma}_{\phi}\}\right],
\end{equation}

where $\Sigma_{\phi}$ is the covariance matrix of of $\pi_{\phi}$ before the parameter update, $\Sigma_{\phi'}$ is the covariance matrix after the parameter update, $k$ is the action dimension, and $\mu_{\phi}$ and $\mu_{\phi'}$ are the means of the policies before and after the parameter update, respectively.
In our implementation, we set the covariance of $\Sigma = \sigma I$ where $\sigma$ is a constant positive value and $I$ is an identity matrix. Then, we can simply the notation as,

\begin{equation}
    D_{KL}(\pi_{\phi}||\pi_{\phi'}) = \dfrac{1}{2} (\mu_{\phi'} - \mu_{\phi})^T{\Sigma}_{\phi'}^{-1}(\mu_{\phi'} - \mu_{\phi}) ,
\end{equation}

\newpage 
\section{Appendix B: Environment}
\label{sec:Environment}

In all environments, the goal of the agent is to reach the desired goal position, regardless of obstacles. The agent constantly receives a scalar reward that is defined as a negative distance from its current position to the desired goal, i.e., $R(s_t, a_t) = -\sqrt{\sum_i |g_i - s_i|^2}$ where $i\in \{0, 1\}$ for 2D and $\{0, 1, 2\}$ for 3D goals. As the simulated robot gets closer to the goal, the reward increases. AntMaze is a maze in a shape of $\supset$ where an agent starts from the bottom left at location (0, 0) and aims to reach the top left goal at location of (0, 16). This experiment tests whether an algorithm can avoid one of the local optima to search for a better policy. Next, AntPush is a maze where a large box is placed in the center of the environment to block the agent from reaching the final goal at location (0, 19). The agent's goal is to move the block so that its path to the goal is cleared out. Similarly, in AntFall, the agent has to push a block into a ditch to cross over the ditch to get to its other end to reach the final goal at location (0, 27, 4.5). This environment extends the navigation task to three dimensions. In both AntPush and AntFall, the greedy algorithm will fail to push the block first, but rather takes a greedy action and become stuck in a spot with decent rewards but not the higher possible reward. Across all experiments, a four-legged Ant is used as a robotic platform. Each leg has two limbs and each limb has a joint. The ant controls all of its joints to walk itself to its desired position. The control input is consisted of these eight inputs. It then observes 31 dimensional states that consists of ant location (3), ant orientation (4), joint angles (8), and all of their velocities, plus time step $t$. The limit of each state and control input are detailed as follows.

The limit of the control input space $\pm[30, 30, 30, 30, 30, 30, 30, 30]$

The limit of the sub-goal space $\pm[10, 10, 0.5, 1, 1, 1, 1, 0.5, 0.3, 0.5, 0.3, 0.5, 0.3, 0.5, 0.3]$

\section{Appendix C: Implementation Details}
\label{sec:Implementation}

We referenced the implementation of HIRO and used the same architecture and hyperparameters.
We adopted the basic TD3 architecture for the deterministic policy in both levels. 
As reported in HIRO, we also modified the layer size of (400, 300) to (300, 300). 
In both layers, we mostly used the same hyperparameters: 
discount factor $\gamma=0.99$, actor learning rate 0.0001, critic learning rate 0.001, soft update targets $\tau=0.005$, 
replay buffer of size 200,000, no gradient clipping, and Gaussian noise with $\sigma=1.0$ for exploration. 
For the reward scaling, 0.1 was used for the high-level and 1.0 for the low-level. The training step and target update interval were set to 1 environment step for low-level and 10 environment steps for high-level. Lastly, the time step $c$ is set to 10 environment steps.






Soft Constraint:
\begin{equation}
    \max_\phi J(\pi^l) = \max_\phi \sum_{t=0}^{T} \mathbb{E}_{(s_t, a_t) \sim \rho^{\pi^l}}  \Bigl[ r (s_t, a_t) - \alpha_{KL} \cdot \rm{KL}(\pi^l_{\phi'} (\cdot | s_t) \| \pi^l_{\phi}(\cdot | s_t))) \Bigr]
\end{equation}

For the soft constraint, we can add a KL divergence loss to the original policy loss when updating the current policy, i.e.,

\begin{equation}
    L'_\phi(\phi) = L_\phi(\phi) + \alpha_{\rm{KL}} \cdot \mathbb{E}_{s_t \sim \rho^{\pi^l}} \Bigl[ \rm{KL}(\pi^l_{\phi^{\prime}} (\cdot | s_t) \| \pi^l_{\phi}(\cdot | s_t))) \Bigr]
\end{equation}

Hard Constraint:
E-Step
\begin{align}
    \max_\phi J(\pi^l) &= \max_\phi \sum_{t=0}^{T} \mathbb{E}_{(s_t, a_t) \sim \rho^{\pi^l}}  \Bigl[ r (s_t, a_t) \Bigr] \\
    &\rm{s.t.}  \quad \mathbb{E}_{s_t \sim \rho^{\pi^l}} \Bigl[ \rm{KL}(\pi^l_{\phi^{\prime}} (\cdot | s_t) \| \pi^l_{\phi}(\cdot | s_t)))\Bigr]  \leq \epsilon
\end{align}

In order to strictly constrain the KL divergence during the policy update, we need to derive an alternative method for optimizing the the policy. We referred to Maximum a Posteriori Policy Optimisation (MPO) algorithm \cite{abdolmaleki2018maximum} that optimizes policy with the family of Expectation Maximization (EM) algorithm.

\begin{algorithm}
\caption{SHIRO Psuedo-Code}
\begin{algorithmic}
    \State Initialize Critic Networks $Q_{\theta^{h_1}}, Q_{\theta^{h_2}}, Q_{\theta^{l_1}}, Q_{\theta^{l_2}}$ with random 
    parameters $\theta^{h_1}, \theta^{h_2}, \theta^{l_1}, \theta^{l_2}$
    \State Initialize Target Networks with parameters $\theta^{h_1 \prime} \gets \theta^{h_1},$ $\theta^{h_2 \prime} \gets \theta^{h_2},$ $\theta^{l_1 \prime} \gets \theta^{l_1},$ $\theta^{l_2 \prime} \gets \theta^{l_2}$
    \State Initialize Actor Networks $\pi^h$ and $\pi^l$ with random parameters $\phi^h, \phi^l$
    \State Initialize Replay Buffers $\beta^h$ and $\beta^l$
    \State Get initial environment state $s_0$ and goal $g$
    \State Temperature parameters $\alpha^h, \alpha^l$
    \For{each environment step $t$}
        \State Select action $a_t \sim \pi^l(s_t, g_t)$ and get new observation and reward $s_{t+1}, r_t$
        \If{step \% c == 0}
            \State Select sub-goal $g_{t+1} \sim \pi^h(s_t, g_t)$
        \Else
            \State $g_{t+1} \gets s_t + g_t - s_{t+1}$
        \EndIf
        \State Store transition tuple $(s_t, g_t, a_t, r_t, s_{t+1}, g_{t+1}, done_t)$ in $\beta^l$
        \If{step \% low\_train\_freq == 0}
            \State sample mini-batch of N transitions from $\beta^l$
            \If{$\alpha^l$}
                \State Entropy $e \sim \pi_{\phi}^l(s_{t+1}, g_{t+1}) $
            \Else
                \State $e = 0$
            \EndIf
            \State $Q_{target}^i \gets r_t + \gamma \cdot (1 - done) \cdot (Q_{\theta^{l_i}}'(s_{t+1}, g_{t_1})) - \alpha e$
            \State $Q_{predict}^i \gets Q_{\theta^{l_i}}(s_t, g_t)$
            \State $\theta^{l_i} \gets \theta^{l_i} - \eta_{\theta}\nabla\mathcal{L}(Q_{target}^i, Q_{predict}^i)$
            \If{step \% low\_update\_delay == 0}
                \State $\phi^l \gets \phi^l - \eta_{\phi}\nabla\mathbb{E}[Q_{\theta^{l_1}}(s_t, g_t) - \alpha e]$
                \State $\theta^{l_i \prime} \gets \tau \theta^{l_i} + (1-\tau)\theta^{l_i \prime}$
                \State $\phi^{l \prime} \gets \tau \phi^l + (1-\tau)\phi^{l \prime}$
                \If {$\alpha^l$}
                    \State $\alpha^l \gets \alpha^l - \eta_{\alpha}\nabla\mathbb{E}[\alpha
                    \pi_{\phi}^l]$
                \EndIf
            \EndIf
        \EndIf
        \If{step \% high\_train\_freq == 0}
            \State Store transition tuple $(s_{t-N:t}, g_{t-N:t}, a_{t-N:t}, R_{t-N:t}, s_{t+1}, done_{t-N:t})$ in $\beta^h$
            \State sample mini-batch of N transitions from $\beta^h$
            \If{$\alpha^h$}
                \State Entropy $e \sim \pi_{\phi}^h(s_{t+1}, g_{t+1}) $
            \Else
                \State $e = 0$
            \EndIf
            \State $Q_{target}^i \gets R_t + \gamma \cdot (1 - done) \cdot (Q_{\theta^{h_i}}'(s_{t+1}, g_{t_1})) - \alpha e$
            \State $Q_{predict}^i \gets Q_{\theta^{h_i}}(s_t, g_t)$
            \State $\theta^{h_i} \gets \theta^{h_i} - \eta_{\theta}\nabla\mathcal{L}(Q_{target}^i, Q_{predict}^i)$
            \If{step \% high\_update\_delay}
                \State $\phi^h \gets \phi^h - \eta_{\phi}\nabla\mathbb{E}[Q_{\theta^{h_1}}(s_t, g_t) - \alpha e]$
                \State $\theta^{h_i \prime} \gets \tau \theta^{h_i} + (1-\tau)\theta^{h_i \prime}$
                \State $\phi^{h \prime} \gets \tau \phi^h + (1-\tau)\phi^{h \prime}$
                \If {$\alpha^h$}
                    \State $\alpha^h \gets \alpha^h - \eta_{\alpha}\nabla\mathbb{E}[\alpha
                    \pi_{\phi}^h]$
                \EndIf
            \EndIf
        \EndIf

    \EndFor
\end{algorithmic}
\end{algorithm}

\newpage 
\section{Appendix D: Additional Analysis and Figures}
\label{sec:Additional_Figures}

We included additional figures for the analysis conducted in \Cref{sec:klAndLowLevelTemp}.

\begin{figure}[h]
    \centering
    \begin{subfigure}[b]{.4\linewidth}
        \includegraphics[width=\linewidth]{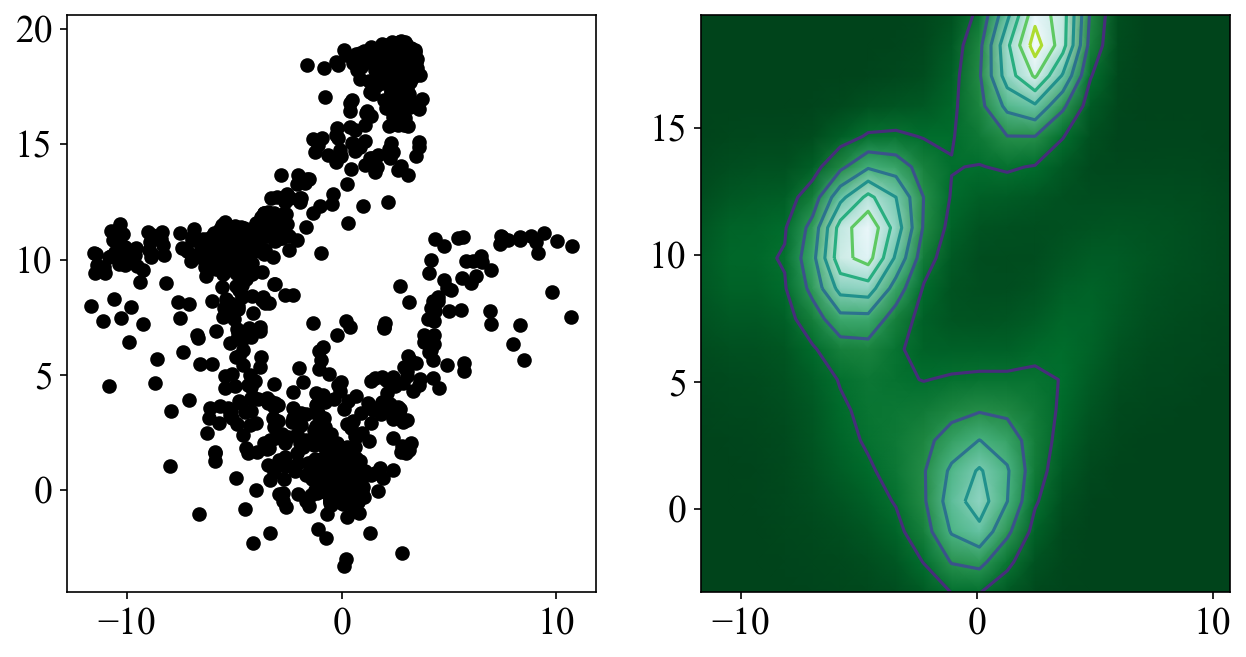}
        \caption{Antpush HIRO}\label{fig:antpush_hiro_project}
    \end{subfigure}
    \begin{subfigure}[b]{.4\linewidth}
        \includegraphics[width=\linewidth]{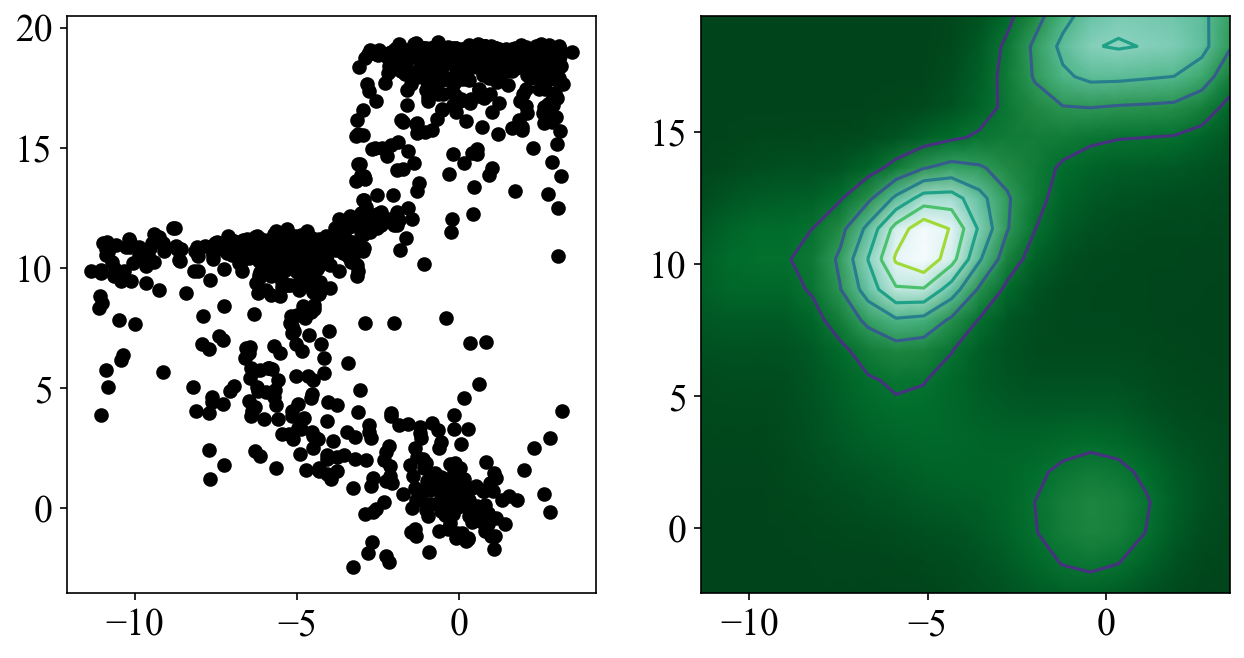}
        \caption{Antpush Learned Entropy SHIRO}\label{fig:antpush_shiro_project}
    \end{subfigure}

    \begin{subfigure}[b]{.4\linewidth}
        \includegraphics[width=\linewidth]{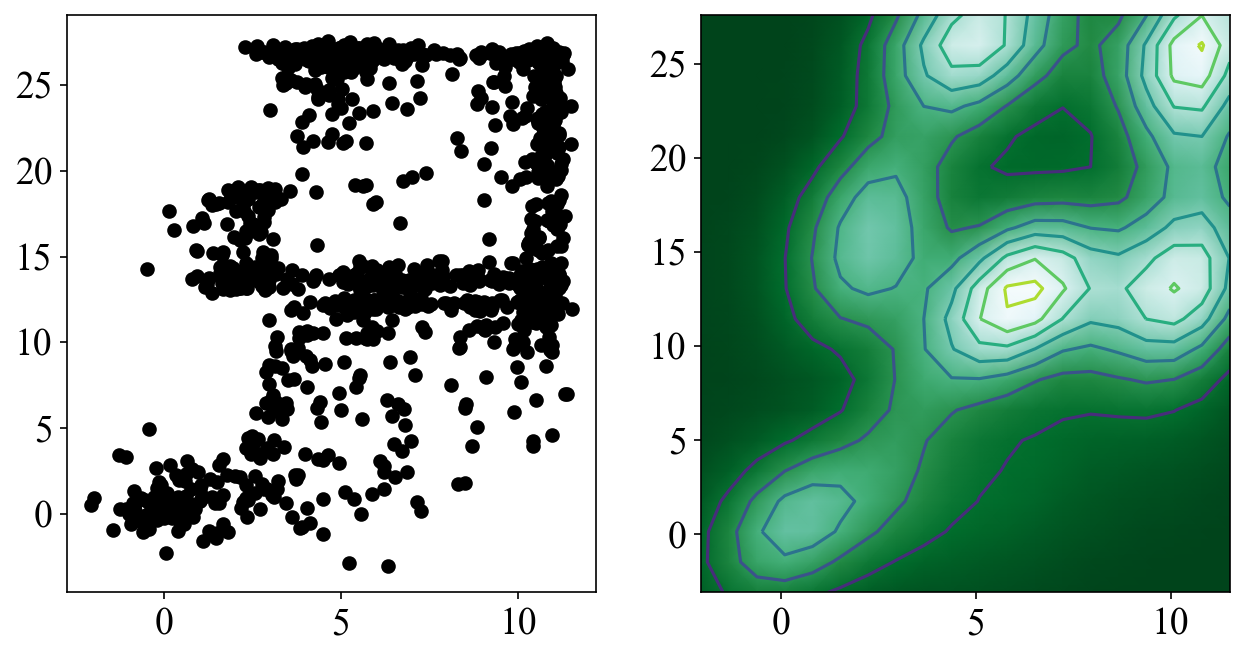}
        \caption{Antfall HIRO}\label{fig:antfall_hiro_project}
    \end{subfigure}
    \begin{subfigure}[b]{.4\linewidth}
        \includegraphics[width=\linewidth]{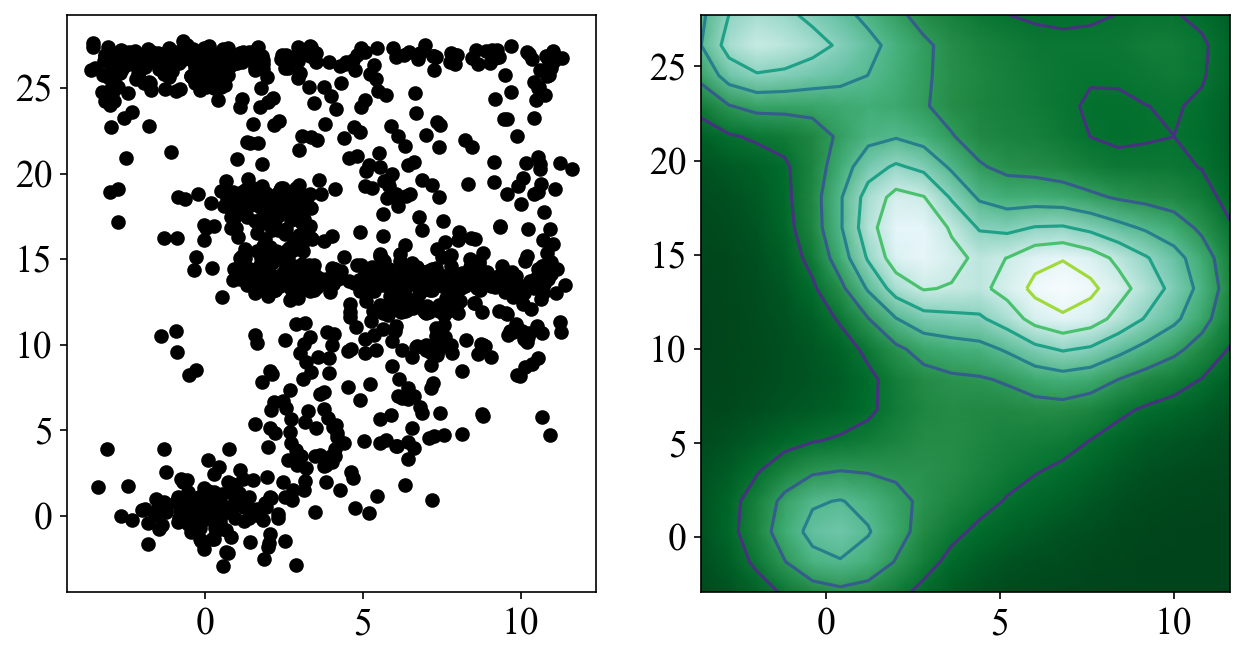}
        \caption{Antfall Learned Entropy SHIRO}\label{fig:antfall_shiro_project}
    \end{subfigure}
    \caption{Analysis of the final positions achieved by an agent in a given environment. In each subfigure, the left plot is simply a scatterplot of the final positions of an agent on the environment specifies underneath. The second plot takes the final position data and displays a contour map, showing where the agent ends up on average across the entire experiment. We find that across all environments tested on, SHIRO HL-Learned explores the environment more thoroughly than HIRO.}
\end{figure}

\begin{figure}[h]
    \centering
    \begin{subfigure}[b]{.3\linewidth}
        \includegraphics[width=\linewidth]{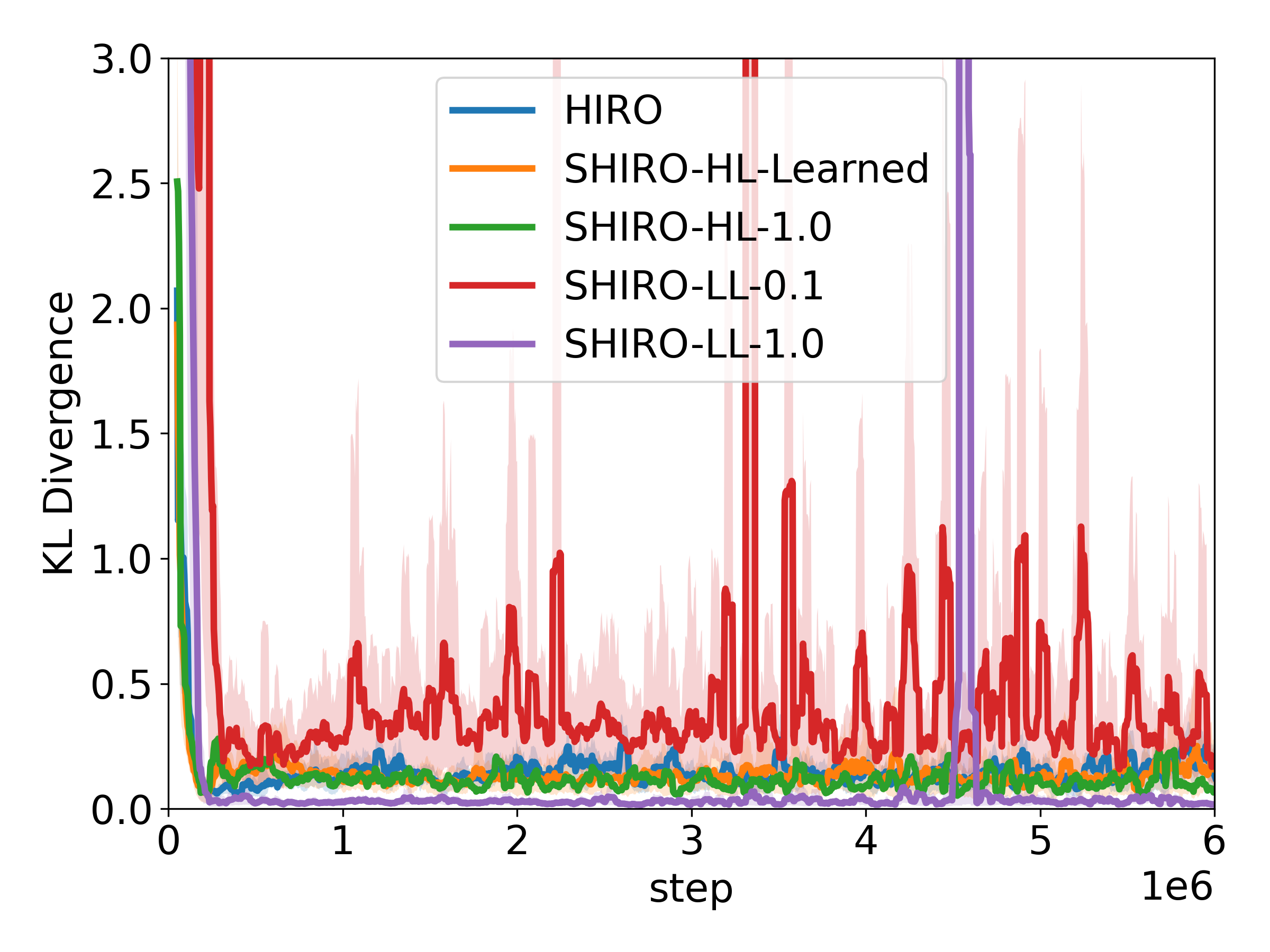}
        \caption{KL Divergence of the low-level policies}
        \label{fig:high_level_temp_a}
    \end{subfigure}
    \begin{subfigure}[b]{.3\linewidth}
        \includegraphics[width=\linewidth]{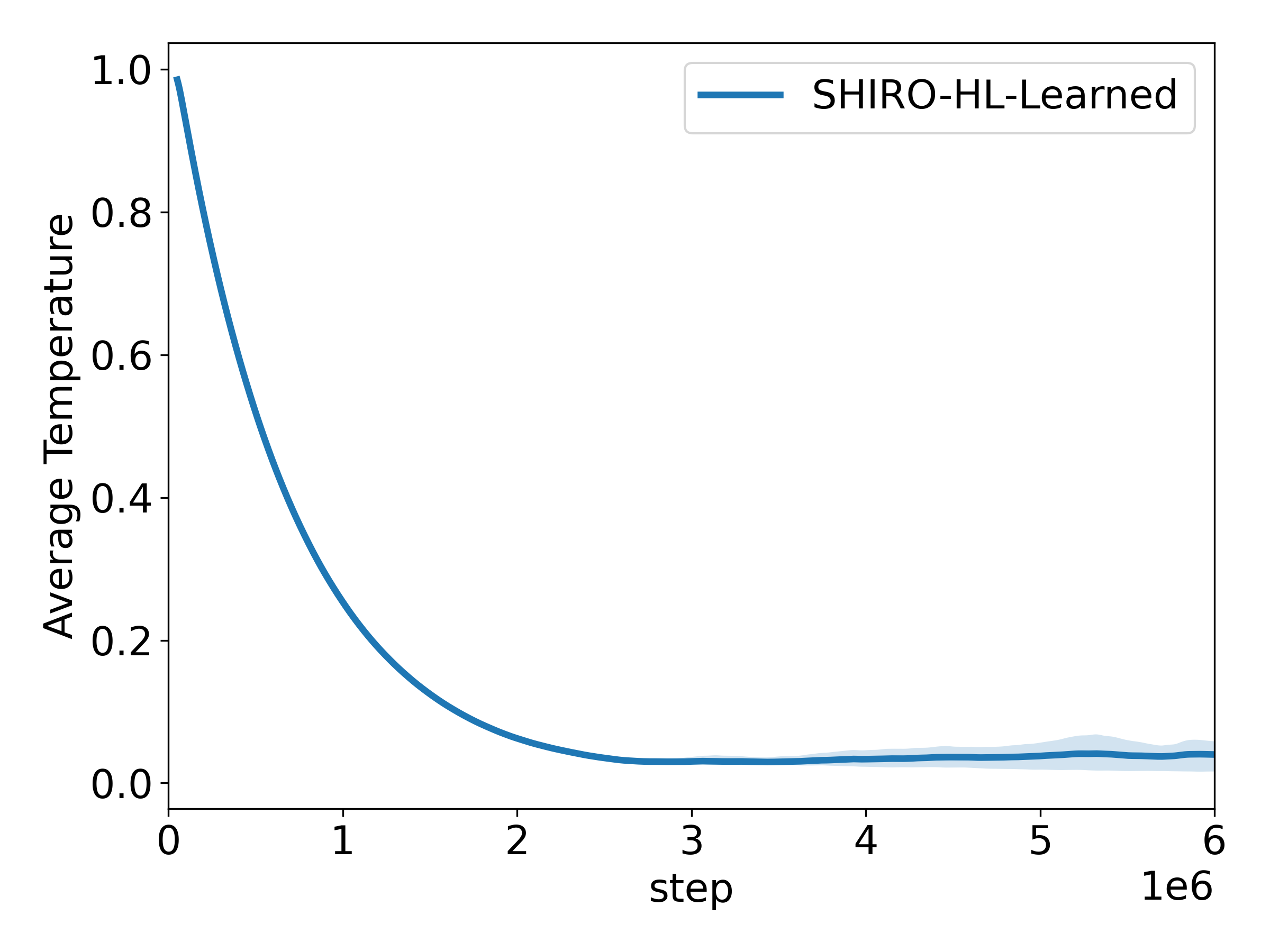}
        \caption{Average Temperature $\alpha$ of the high-level policy}
        \label{fig:high_level_temp_b}
    \end{subfigure}
    \begin{subfigure}[b]{.3\linewidth}
        \includegraphics[width=\linewidth]{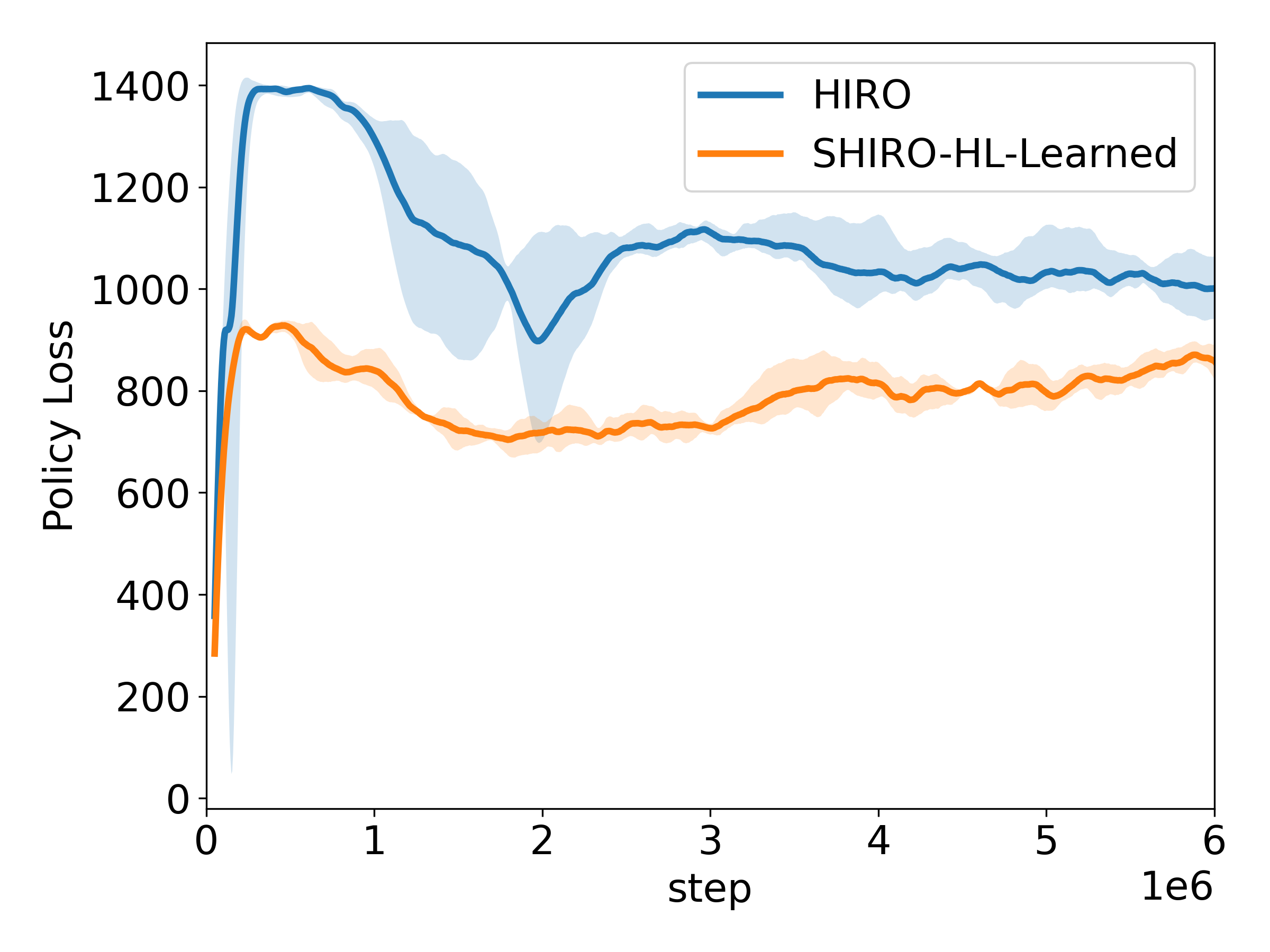}
        \caption{Policy loss of the low-level policies}\label{fig:policy_loss}
    \end{subfigure}

    \begin{subfigure}[b]{.3\linewidth}
        \includegraphics[width=\linewidth]{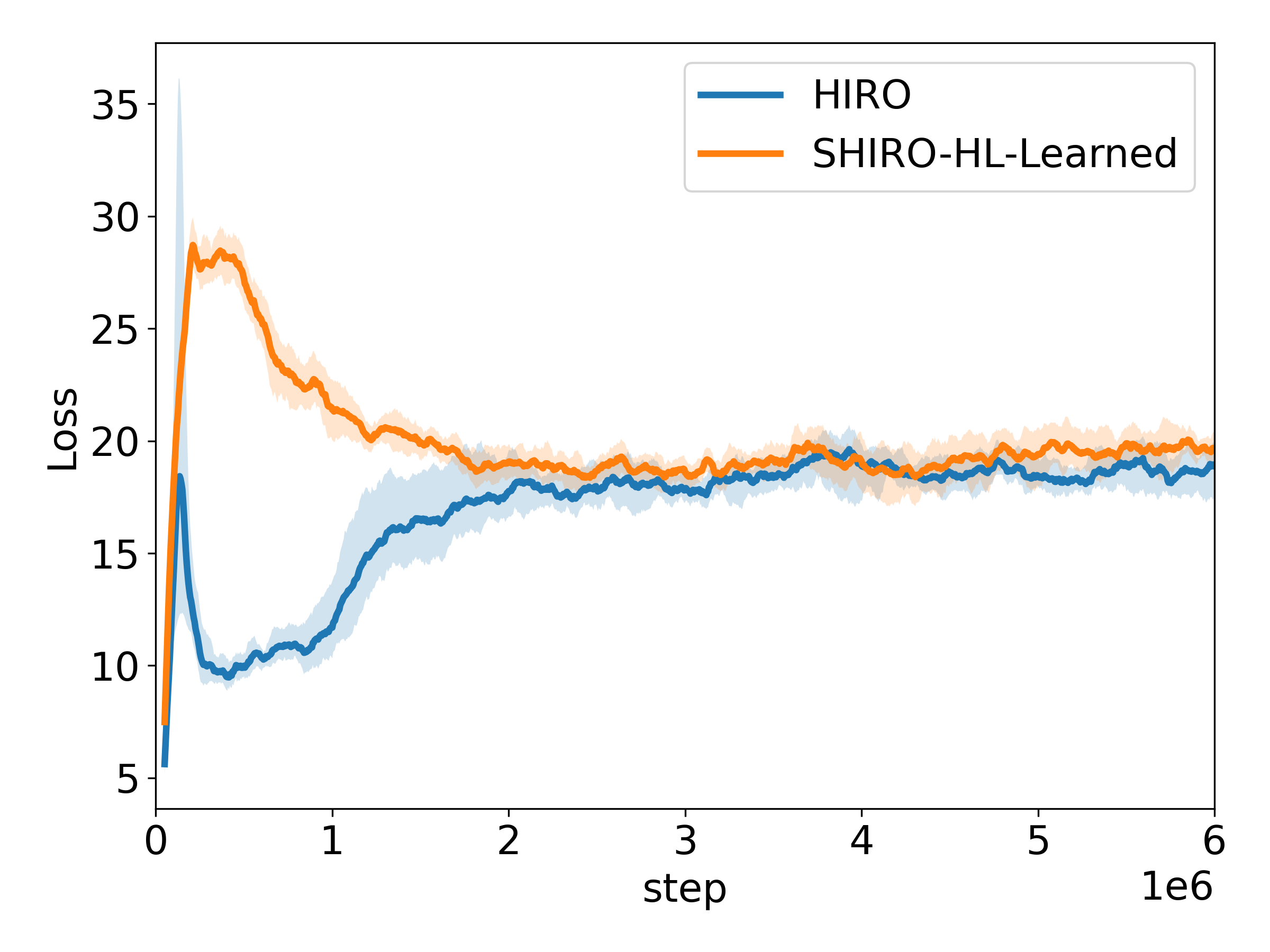}
        \caption{$Q$ function loss of the high-level policies}\label{fig:qfunc_loss}
    \end{subfigure}
    \begin{subfigure}[b]{.3\linewidth}
        \includegraphics[width=\linewidth]{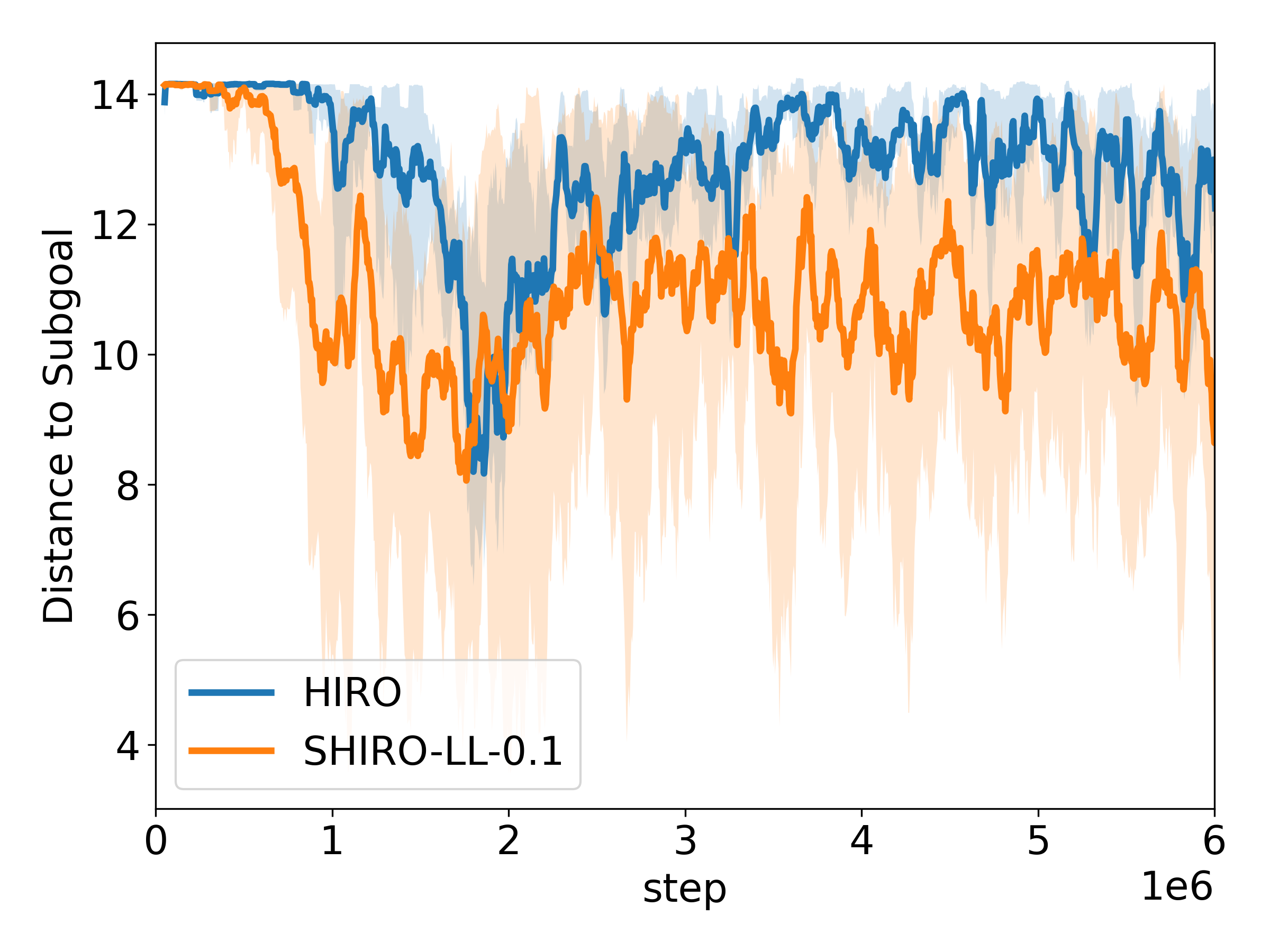}
        \caption{Distance to sub-goals (HIRO vs. SHIRO-LL with $\alpha=0.1$)}\label{fig:hiro_vs_ll}
    \end{subfigure}
    \begin{subfigure}[b]{.3\linewidth}
        \includegraphics[width=\linewidth]{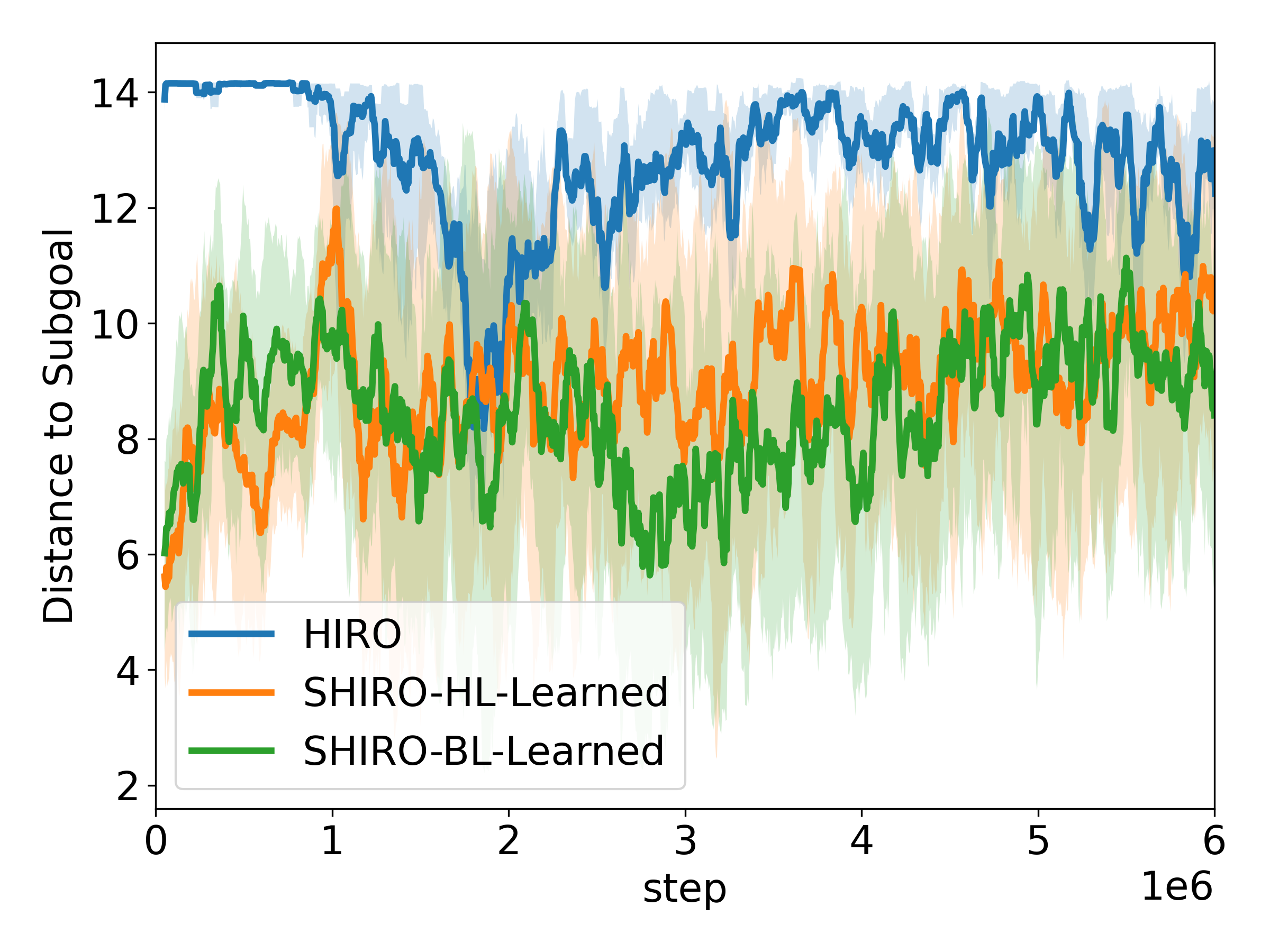}
        \caption{Distance to sub-goals (HIRO vs. SHIRO-HL/BL-Learned)}\label{fig:agent_diff}
    \end{subfigure}
    \caption{Additional figures for the analysis of the entropy effects.}
\end{figure}

\end{document}


\twocolumn[
\icmltitle{Appendix}



\icmlsetsymbol{equal}{*}

\begin{icmlauthorlist}
\icmlauthor{Kandai Watanabe}{cu}
\icmlauthor{Matthew Strong}{cu}
\icmlauthor{Omer Eldar}{cu}
\icmlauthor{Advisor}{cu}
\end{icmlauthorlist}

\icmlaffiliation{cu}{Department of Computer Science, University of Colorado Boulder, Colorado, USA}

\icmlcorrespondingauthor{Kandai Watanabe}{kandai.watanabe@colorado.edu}

\icmlkeywords{Off-Policy, Hierarchical, Reinforcement Learning, Maximum Entropy}

\vskip 0.3in
]


\printAffiliationsAndNotice{}  

